%% file: main.tex
\documentclass[twoside]{article}
\pdfoutput=1
\usepackage[accepted]{aistats2018}

\input{math_definition.tex}

\newtheorem{innercustomthm}{Lemma\,}
\newenvironment{customlemma}[1]
  {\renewcommand\theinnercustomthm{#1}\innercustomthm}
  {\endinnercustomthm}

\newlength{\textfloatsepsave} 

\begin{document}

\twocolumn[

\aistatstitle{Batched Large-scale Bayesian Optimization in High-dimensional Spaces} %

\aistatsauthor{ Zi Wang \And Clement Gehring \And Pushmeet Kohli \And Stefanie Jegelka}

\aistatsaddress{ MIT CSAIL \And  MIT CSAIL \And DeepMind  \And MIT CSAIL} ]

\begin{abstract}
\input{abstract}
\end{abstract}

\input{intro}

\input{related}

\input{background}
\input{approach2}

\input{connection}

\input{exp}

\input{rover}

\input{conclu}
\section*{Acknowledgements}
We gratefully acknowledge support from NSF CAREER award 1553284, NSF grants 1420927 and 1523767, from ONR grant N00014-14-1-0486, and from ARO grant W911NF1410433.  Any opinions, findings, and conclusions or recommendations expressed in this material are those of the authors and do not necessarily reflect the views of our sponsors. 

{
\bibliographystyle{plainnat}
\bibliography{refs}
}
\newpage~\newpage

\appendix
\input{sup_method}

\input{sup_exp}
\input{sup_disc}

\end{document}

%% file: math_definition.tex
\usepackage{tikz}
\usepackage{mathtools}
\usetikzlibrary{fit,positioning}
\usepackage{diagbox}
\usepackage{bbm}
\usepackage{times}
\usepackage{graphicx} 
\usepackage{subfigure}

\usepackage[round]{natbib}

\usepackage{algorithmicx}
\usepackage{algorithm}
\usepackage{algpseudocode}
\algnewcommand{\LineComment}[1]{\Statex \(\triangleright\) #1}
\algblock{ParFor}{EndParFor}
\algnewcommand\algorithmicparfor{\textbf{parfor}}
\algnewcommand\algorithmicpardo{\textbf{do}}
\algnewcommand\algorithmicendparfor{\textbf{end\ parfor}}
\algrenewtext{ParFor}[1]{\algorithmicparfor\ #1\ \algorithmicpardo}
\algrenewtext{EndParFor}{\algorithmicendparfor}

\usepackage{graphicx}
\usepackage{amsmath,amsthm,amssymb,bm} 
\usepackage{amsfonts}
\usepackage{mathrsfs}
\usepackage{subfigure}
\usepackage{xspace}
\usepackage{array}
\usepackage{enumerate}
\usepackage{stmaryrd}
\usepackage{appendix}
\usepackage{wrapfig}
\numberwithin{equation}{section}

   \newcommand{\cd}{\mathcal{D}}   \newcommand{\cg}{\mathcal{G}}         \newcommand{\cp}{\mathcal{P}}       \newcommand{\cx}{\mathcal{X}}  

\theoremstyle{plain}

\theoremstyle{definition}

\theoremstyle{remark}

\usepackage{color}

\usepackage{microtype}      
\usepackage{nicefrac}       
\usepackage{booktabs}       
\usepackage{hyperref}
\usepackage{graphicx} 
\usepackage{booktabs} 
\usepackage{amssymb}
\usepackage{times}
\usepackage{graphicx}
\usepackage{color}
\usepackage{url}
\usepackage{bbm}
\usepackage{multicol}

\providecommand{\dif}{\mathop{}\!\mathrm d}
\providecommand{\Ex}{\mathbb E}
\providecommand{\hide}[1]{}

\usepackage{amsmath,amsfonts}
\usepackage{amsopn,amssymb,amsthm,thmtools,thm-restate}

\theoremstyle{plain}
\newtheorem{thm}{Theorem}[section]
\newtheorem{lem}[thm]{Lemma}

\theoremstyle{definition}

\theoremstyle{remark}

\usepackage{bm} 
\usepackage{algorithm}
\newcommand{\vct}[1]{\boldsymbol{#1}} 
\newcommand{\mat}[1]{\boldsymbol{#1}} 

\newcommand{\field}[1]{\mathbb{#1}}
\newcommand{\R}{\field{R}} 
\newcommand{\T}{^{\textrm T}} 

\newcommand{\ProbOpr}[1]{\mathbb{#1}}

\newcommand{\expect}[2]{%
\ifthenelse{\equal{#2}{}}{\ProbOpr{E}_{#1}}
{\ifthenelse{\equal{#1}{}}{\ProbOpr{E}\left[#2\right]}{\ProbOpr{E}_{#1}\left[#2\right]}}} 
\newcommand{\var}[2]{%
\ifthenelse{\equal{#2}{}}{\ProbOpr{VAR}_{#1}}
{\ifthenelse{\equal{#1}{}}{\ProbOpr{VAR}\left[#2\right]}{\ProbOpr{VAR}_{#1}\left[#2\right]}}} 

\DeclareMathOperator{\argmax}{arg\,max}
\DeclareMathOperator{\argmin}{arg\,min}

\newcommand{\vpp}{\vct{\phi}}
\newcommand{\mphi}{\mat{\Phi}}
\newcommand{\vx}{{\vct{x}}}
\newcommand{\vy}{\vct{y}}

\newcommand{\vb}{\vct{b}}

\newcommand{\kk}{\kappa}
\newcommand{\vkk}{\vct{\kappa}}

\newcommand{\mI}{\mat{I}}
\newcommand{\mK}{\mat{K}}

\newcommand{\mSigma}{\mat{\Sigma}}



%% file: abstract.tex
Bayesian optimization (BO) has become an effective approach for black-box function optimization problems when function evaluations are expensive and the optimum can be achieved within a relatively small number of queries. However, many cases, such as the ones with high-dimensional inputs, may require a much larger number of observations for optimization. Despite an abundance of observations thanks to parallel experiments, current BO techniques have been limited to merely a few thousand observations. In this paper, we propose \emph{ensemble Bayesian optimization} (EBO) to address three current challenges in BO \emph{simultaneously}: (1)  large-scale observations; (2) high dimensional input spaces; and (3) selections of batch queries that balance quality and diversity. The key idea of EBO is to operate on an ensemble of additive Gaussian process models, each of which possesses a randomized strategy to divide and conquer.  We show unprecedented, previously impossible results of scaling up BO to tens of thousands of observations within minutes of computation. 

%% file: intro.tex
\section{Introduction}
Global optimization of black-box and non-convex functions is an important component of modern machine learning. From optimizing hyperparameters in deep models to solving inverse  problems encountered in computer vision and policy search for reinforcement learning, these optimization problems have many important applications in machine learning and its allied disciplines. In the past decade, Bayesian optimization has become a popular approach for global optimization of non-convex functions that are expensive to evaluate. Recent work addresses better query strategies~\citep{kushner1964,mockus1974, srinivas2012information, hennig2012,hernandez2014predictive,wang2016est,kawaguchi2015bayesian}, techniques for batch queries~\citep{desautels2014parallelizing,gonzalez2016batch}, and algorithms for high dimensional problems~\citep{wang2016bayesian,kandasamy2015high}. 

Despite the above-mentioned successes, Bayesian optimization remains somewhat impractical, since it is typically coupled with expensive function estimators (Gaussian processes) and non-convex acquisition functions that are hard to optimize in high dimensions and sometimes expensive to evaluate. To alleviate these difficulties, recent work explored the use of random feature approximations~\citep{snoek2015scalable, lakshminarayanan2016mondrian} and sparse Gaussian processes~\citep{mcintire2016sparse}, but, while improving scalability, these methods still suffer from misestimation of confidence bounds (an essential part of the acquisition functions), and expensive or inaccurate Gaussian process (GP) hyperparameter inference. Indeed, to the best of our knowledge, Bayesian optimization is typically limited to a few thousand evaluations~\citep{lakshminarayanan2016mondrian}. Yet, reliable search and estimation for complex functions in very high-dimensional spaces may well require more evaluations. 
With the increasing availability of parallel computing resources, large number of function evaluations are possible if the underlying approach can leverage the parallelism.
Comparing to the millions of evaluations possible (and needed) with \emph{local} methods like stochastic gradient descent, the scalability of \emph{global} Bayesian optimization leaves large room for desirable progress. In particular, the lack of scalable uncertainty estimates to guide the search is a major roadblock for huge-scale Bayesian optimization.

In this paper, we propose ensemble Bayesian optimization (EBO), a global optimization method targeted to high dimensional, large scale parameter search problems whose queries are parallelizable. Such problems are abundant in hyper and control parameter optimization in machine learning and robotics~\citep{calandra2017bayesian,snoek2012practical}. EBO relies on two main ideas that are implemented at multiple levels: (1) we use efficient partition-based function approximators (across both data \emph{and} features) that simplify and accelerate search and optimization; (2) we enhance the expressive power of these approximators by using ensembles and a stochastic approach.
We maintain an evolving (posterior) distribution over the (infinite) ensemble and, in each iteration, draw one member to perform search and estimation. 

In particular, we use a new combination of three types of partition-based approximations: (1-2) For improved GP estimation, we propose a novel \emph{hierarchical} additive GP model based on tile coding (a.k.a. random binning or Mondrian forest features). We learn a posterior distribution over kernel width and the additive structure; here, Gibbs sampling prevents overfitting. (3) To accelerate the sampler, which depends on the likelihood of the observations, we use an efficient, randomized block approximation of the Gram matrix based on a Mondrian process. Sampling and query selection can then be parallelized across blocks, further accelerating the algorithm.

As a whole, this combination of simple, tractable structure with ensemble learning and randomization improves efficiency, uncertainty estimates and optimization. Moreover, we show that our realization of these ideas offers an alternative explanation for global optimization heuristics that have been popular in other communities, indicating possible directions for further theoretical analysis. 
Our empirical results demonstrate that EBO can speed up the posterior inference by 2-3 orders of magnitude (400 times in one experiment) compared to the state-of-the-art, without sacrificing quality. Furthermore, we demonstrate the ability of EBO to handle sample-intensive hard optimization problems by applying it to real-world problems with tens of thousands of observations.

%% file: related.tex
\paragraph{Related Work}
There has been a series of works addressing the three big challenges in BO: selecting batch evaluations \citep{contal2013parallel, desautels2014parallelizing,gonzalez2016batch,wang2017batched, daxberger2017distributed}, high-dimensional input spaces \citep{wang2016bayesian,djolonga2013high,li2016high,kandasamy2015high,wang2017batched,wang2017maxvalue}, and scalability \citep{snoek2015scalable, lakshminarayanan2016mondrian, mcintire2016sparse}. Although these three problems tend to co-occur, this paper is the first (to the best of our knowledge) to address all three challenges jointly in one framework.

Most closely related to parts of this paper is \citep{wang2017batched}, but our algorithm significantly improves on that work in terms of scalability (see Sec.~\ref{ssec:exp:scalability} for an empirical comparison), and has fundamental technical differences. First, the Gibbs sampler by~\cite{wang2017batched} only learns the additive structure but not the kernel parameters, while our sampler jointly learns both of them. Second, our proposed algorithm partitions the input space for scalability and parallel inference. We achieve this by a Mondrian forest. Third, as a result, our method automatically generates batch queries, while the other work needs an explicit batch strategy.

Other parts of our framework are inspired by the Mondrian forest~\citep{lakshminarayanan2016mondrian}, which partitions the input space via a Mondrian tree and aggregates trees into a forest. The closely related Mondrian kernels \citep{balog2016mondrian} use random features derived from Mondrian forests to construct a kernel. Such a kernel, in fact, approximates a Laplace kernel. In fact, Mondrian forest features can be considered a special case of the popular tile coding features widely used in reinforcement learning~\citep{sutton1998reinforcement,albus1975new}. %
\cite{lakshminarayanan2016mondrian} showed that, in low-dimensional settings, Mondrian forest kernels scale better than the regular GP and achieve good uncertainty estimates in many low-dimensional problems.

Besides Mondrian forests, there is a rich literature on sparse GP methods to address the scalability of GP regression \citep{seeger2003fast, snelson2006sparse, titsias2009variational, hensman2013gaussian}. However, these methods are mostly only shown to be useful when the input dimension is low and there exist redundant data points, so that inducing points can be selected to emulate the original posterior GP well. However, data redundancy is usually not the case in high-dimensional Bayesian optimization. Recent applications of sparse GPs in BO~\citep{mcintire2016sparse} only consider experiments with less than 80 function evaluations in BO and do not show results on large scale %
observations. Another approach to tackle large scale GPs distributes the computation via local experts~\citep{deisenroth2015distributed}. However, this is not very suitable for the acquisition function optimization needed in Bayesian optimization, since every valid prediction needs to synchronize the predictions from all the local experts. Our paper is also related to~\citet{gramacy2008bayesian}. While \citet{gramacy2008bayesian} focuses on modeling non-stationary functions with treed partitions, our work integrates tree structures and Bayesian optimization in a novel way.

%% file: background.tex
\section{Background and Challenges}
Consider a simple but high-dimensional search space $\cx = [0, R]^D\subseteq \mathbb R^{D}$. We aim to find a maximizer $x^* \in \arg\max_{x\in\cx} f(x)$ of a black-box function $f:\cx\to\mathbb{R}$.

\paragraph{Gaussian processes.}
Gaussian processes (GPs) are popular priors for modeling the function $f$ in Bayesian optimization. They define distributions 
over functions where any finite set of function values has a multivariate Gaussian distribution. A Gaussian process $\cg\cp(\mu,\kk)$ is fully specified by a mean function $\mu(\cdot)$ and covariance (kernel) function $\kk(\cdot,\cdot)$. 
Let $f$ be a function sampled from $\cg\cp(0,\kk)$. 
Given observations $\mathcal D_{n} = \{(\vx_t, y_t)\}_{t=1}^n$ where $y_t\sim\mathcal N(f(\vx_t),\sigma)$, we obtain the posterior mean and variance of the function as
\begin{align}\label{eq:gppost}
\mu_{n}(\vx) &= \vkk_n(\vx)\T(\mK_n+\sigma^2\mI)^{-1}\vy_n, \\
\sigma^2_{n}(\vx) &= \kk(\vx,\vx) - \vkk_n(\vx)\T(\mK_n+\sigma^2\mI)^{-1} \vkk_n(\vx)
\end{align}
via the kernel matrix $\mK_n =\left[\kk(\vx_i,\vx_j)\right]_{\vx_i,\vx_j\in \mathcal D_n}$ and $\vkk_n(\vx) = [\kk(\vx_i,\vx)]_{\vx_i\in D_n}$~\citep{rasmussen2006gaussian}.
 The log data likelihood for $\mathcal D_{n}$ is given by
\begin{align} \label{eq:gploglike}
&\log p(\mathcal D_n)   = -\frac12 \vy_n\T (\mK_n+\sigma^2 \mI)^{-1}\vy_n \nonumber\\
 & \;\;\; -\frac12\log |\mK_n+\sigma^2 \mI| - \frac{n}{2}\log 2\pi.
\end{align}
While GPs provide flexible, broadly applicable function estimators, the $O(n^3)$ computation of the inverse $(\mK_n+\sigma^2\mI)^{-1}$ and determinant $|\mK_n+\sigma^2 \mI|$ can become major bottlenecks as $n$ grows, for both posterior function value predictions and data likelihood estimation.

\paragraph{Additive structure.} \label{ssec:additive}
To reduce the complexity of the vanilla GP, we assume a latent decomposition of the input dimensions $[D]=\{1,\ldots,D\}$ into disjoint subspaces, namely, $\bigcup_{m=1}^M A_m = [D]$ and $A_i\cap A_j = \emptyset$ for all $i\ne j$, $i,j\in[M]$. As a result, the function $f$ decomposes as $f(x) = \sum_{m\in[M]} f_m(x^{A_m})$~\citep{kandasamy2015high}. If each component $f_m$ is drawn independently from $\cg\cp(\mu^{(m)},\kk^{(m)})$ for all $m\in[M]$, the resulting $f$ will also be a sample from a GP: $f\sim\cg\cp(\mu, \kk)$, with $\mu(x) = \sum_{m\in[M]}\mu_m(x^{A_m}), \;\;
\kk(x, x') = \sum_{m\in[M]}\kk^{(m)}(x^{A_m}, {x'}^{A_m})$. 

The additive structure reduces sample complexity and helps BO to search more efficiently and effectively since the acquisition function can be optimized component-wise. But it remains challenging to learn a good decomposition structure $\{A_m\}$. Recently, \cite{wang2017batched} proposed learning via Gibbs sampling. This sampler takes hours 
for merely a few hundred points, because it needs a vast number of expensive data likelihood computations.

\paragraph{Random features.} It is possible use random features~\citep{rahimi2007random} to approximate the GP kernel and alleviate the $O(n^3)$ computation in Eq.~\eqref{eq:gppost} and Eq.~\eqref{eq:gploglike}. Let $\vpp:\cx\mapsto \R^{D_R}$ be the (scaled) random feature operator and $\mphi_n = [\vpp(\vx_1), \cdots, \vpp(\vx_n)]\T \in \R^{n\times D_R}$. The GP posterior mean and variance can be written as%
\begin{align}
\mu_n(\vx) &= \sigma^{-2} \vpp(\vx)\T\mSigma_n \mphi_n\T \vy_n, \\
\sigma_n^2(x) &= \vpp(\vx)\T \mSigma_n \vpp(\vx),
\end{align}
where $\mSigma_n  = (\mphi_n\T \mphi_n \sigma^{-2} +\mI)^{-1}$.
By the Woodbury matrix identity and the matrix determinant lemma, the log data likelihood becomes 
\begin{align} \label{eq:gploglike2}
 & \log p(\mathcal D_n)= \frac{\sigma^{-4}}{2} \vy_n\T\mphi_n \mSigma_n\mphi_n\T\vy_n \nonumber\\
 & \;\;\;-\frac{1}{2}\log |\mSigma_n^{-1}| -\frac{\sigma^{-2}}{2} \vy_n\T\vy_n- \frac{n}{2}\log 2\pi\sigma^2.
\end{align}
The number of random features necessary to approximate the GP well in general increases with the number of observations~\citep{rudi2016generalization}. Hence, for large-scale observations, we cannot expect to solely use a fixed number of features. Moreover, learning hyperparameters for random features is expensive: for Fourier features, the computation of Eq.~\eqref{eq:gploglike2} means re-computing the features, plus $O(D_R^3)$ for the inverse and determinant. With Mondrian features~\citep{lakshminarayanan2016mondrian}, we can learn the kernel width efficiently by adding more Mondrian blocks, but this procedure is not well compatible with learning additive structure, since the whole structure of the sampled Mondrian features will change. In addition, we typically need a forest of trees for a good approximation.
 \begin{figure*}[t]
  \centering
   \includegraphics[width=.85\textwidth]{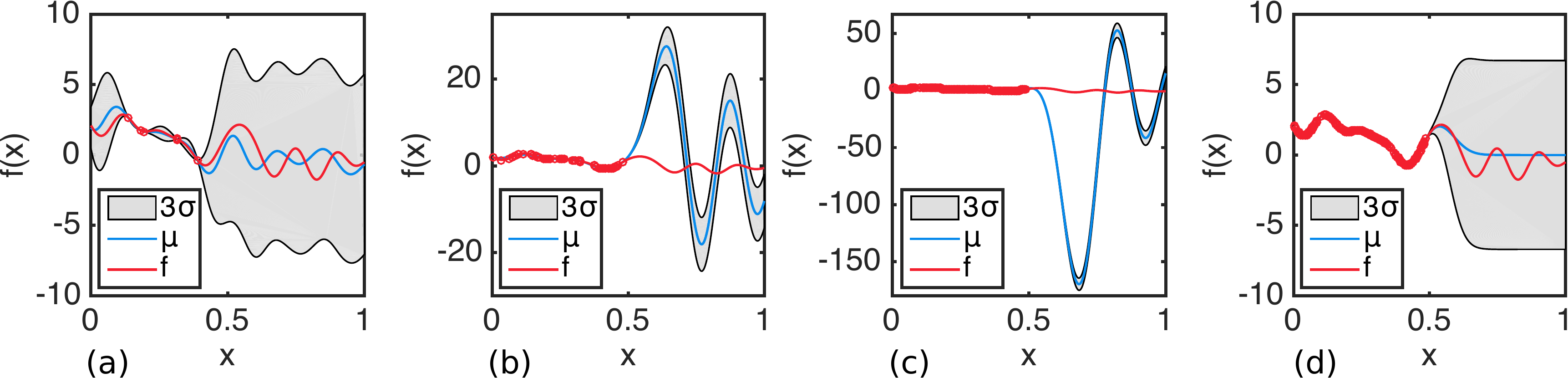}
  \caption{We use 1000 Fourier features to approximate a 1D GP with a squared exponential kernel. The observations are samples from a function $f$ (red line) drawn from the GP with zero mean in the range $[-10,0.5]$. (a) Given 100 sampled observations (red circles), the Fourier features lead to reasonable confidence bounds. (b) Given 1000 sampled observations (red circles), the quality of the variance estimates degrades. (c) With additional samples (5000 observations), the problem is exacerbated. The scale of the variance predictions relative to the mean prediction is very small. (d) For comparison, the proper predictions of the original full GP conditioned on the same 5000 observations as (c). Variance starvation becomes a serious problem for random features when the size of data is close to or larger than the size of the features.}
  \label{fig:badvar}
  \end{figure*}
\paragraph{Tile coding.} 
Tile coding \citep{sutton1998reinforcement,albus1975new} is a $k$-hot encoding widely used in reinforcement learning as an efficient set of non-linear features. In its simplest form, tile coding is defined by $k$ partitions, referred to as layers. An encoded data point becomes a binary vector with a non-zero entry for each bin containing the data point. There exists methods for sampling random partitions that allow to approximate various kernels, such as the `hat' kernel \citep{rahimi2007random}, making tile coding well suited for our purposes.

\paragraph{Variance starvation.}
\label{sec:var}
It is probably not surprising that using finite random features 
to learn the function \emph{distribution} will result in a loss in accuracy \citep{forster2005notice}. 
For example, we observed that, while the mean predictions are preserved reasonably well around regions where we have observations, both mean and confidence bound predictions can become very bad in regions where we do not have observations, once there are more observations than features. We refer to this underestimation of variance scale compared to mean scale, illustrated in Fig.~\ref{fig:badvar}, as \emph{variance starvation}. 

%% file: approach2.tex
\section{Ensemble Bayesian Optimization}
\label{sec:ebo}

Next, we describe an approach that scales Bayesian Optimization when parallel computing resources are available. We name our approach, outlined in Alg.\ref{alg:ebo}, \emph{Ensemble Bayesian optimization (EBO)}. At a high level, EBO uses a (stochastic) series of Mondrian trees to partition the input space, learn the kernel parameters of a GP locally, and aggregate these parameters. Our forest hence spans across BO iterations. 

In the $t$-th iteration of EBO in Alg.~\ref{alg:ebo}, we use a Mondrian process to randomly partition the search space into $J$ parts (line~\ref{mondrian}), where $J$ can be dependent on the size of the observations $\cd_{t-1}$. For the $j$-th partition, we have a subset  $\cd_{t-1}^j$ of observations. From those observations, we learn a local GP with random tile coding \emph{and} additive structure, via Gibbs sampling (line~\ref{gibbs}). For conciseness, we refer to such GPs as TileGPs.
The probabilistic tile coding can be replaced by a Mondrian grid that approximates a Laplace kernel~\citep{balog2015mondrian}. Once a TileGP is learned locally, we can run BO with the acquisition function $\eta$ in each partition to generate a candidate set of points, and, from those, select a batch that is both informative (high-quality) and diverse (line~\ref{filter}).
\begin{algorithm}[H]
  \caption{Ensemble Bayesian Optimization (EBO)}\label{alg:ebo}
    \begin{small}
  \begin{algorithmic}[1]
    \Function{EBO\,}{$f, \cd_0$}
      \State Initialize $z, k$
      \For{$t = 1,\cdots, T $}
      \State $\{\cx_j\}_{j=1}^J\gets$\textsc{Mondrian}($[0,R]^D, z, k, J$)\label{mondrian}
      \ParFor{$j = 1,\cdots, J$}
      \State $z^j, k^j\gets\textsc{GibbsSampling}(z,k \mid \cd^j_{t-1})$\label{gibbs}
      \State $\eta^j_{t-1}(\cdot)\gets$\textsc{Acquisition\,}($\cd_{t-1}^{j}, z^j, k^j$) \label{acfun}
      \State $\{A_m\}_{m=1}^M\gets\textsc{Decomposition}(z^j)$
      \For{$m=1,\cdots,M$} \label{decomp}
      \State $\vx_{tj}^{A_m} \gets \argmax_{\vx\in\cx^{A_m}_j} \eta^j_{t-1}(\vx)$ \label{opt_acfun}
      \EndFor
      \EndParFor
      \State $z\gets\textsc{Sync}(\{z^j\}_{j=1}^J), \;\; k\gets\textsc{Sync}(\{k^j\}_{j=1}^J)$ \label{sync}
      \State $\{\vx_{tb}\}_{b=1}^B\gets$ \textsc{Filter\,}($\{\vx_{tj}\}_{j=1}^J \mid z, k$) \label{filter}
      \ParFor{$b=1,\cdots,B$}
      \State $y_{tb}\gets f(\vx_{tb}) $\label{eval}
      \EndParFor
      \State $\mathfrak \cd_t \gets \cd_{t-1}\cup \{\vx_{tb},y_{tb}\}_{b=1}^B$
      \EndFor
      \EndFunction

  \end{algorithmic}
        \end{small}
\end{algorithm}

Since, in each iteration, we draw an input space partition and update the kernel width and the additive structure, the algorithm may be viewed as implicitly and stochastically running BO on an ensemble of GP models. In the following, we describe our model and the procedures of Alg.~\ref{alg:ebo} in detail. In the Appendix, we show an illustration how EBO optimizes a 2D function. 
\subsection{Partitioning the input space via a Mondrian process}
When faced with a ``big'' problem, a natural idea is to divide and conquer. For large scale Bayesian optimization, the question is how to divide without losing the valuable local information that gives good uncertainty measures. In EBO, we use a Mondrian process to divide the input space and the observed data, so that nearby data points remain together in one partition, preserving locality\footnote{We include the algorithm for input space partitioning in the appendix.}. The Mondrian process uses axis-aligned cuts to divide the input space $[0,R]^D$ into a set of partitions $\{\cx_j\}_{j=0}^J$ where $\cup_j \cx_j = [0,R]^D$ and $\cx_i \cap \cx_j =\emptyset, \forall i\neq j$. Each partition $\cx_j$ can be conveniently described by a hyperrectangle $[l_1^j, h_1^j]\times\cdots \times [l_D^j, h_D^j]$, which facilitates the efficient use of tile coding and Mondrian grids in a TileGP. In the next section, we define a TileGP and introduce how its parameters are learned. %

\subsection{Learning a local TileGP via Gibbs sampling}
\label{ssec:gibbs}

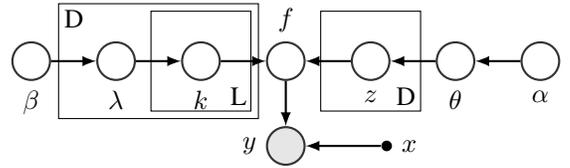
\begin{figure}
\label{fig:addgpgraph}
\centering
\begin{tikzpicture}
\tikzstyle{main}=[circle, minimum size = 5mm, thick, draw =black!80, node distance = 6mm]
\tikzstyle{para}=[circle, minimum size = 4pt, inner sep=0pt]
\tikzstyle{connect}=[-latex, thick]
\tikzstyle{box}=[rectangle, draw=black!100]
  \node[main] (beta) [label=below:$\beta$] { };
    \node[main] (lambda) [right=of beta,label=below:$\lambda$] { };
    \node[main] (k) [right=of lambda,label=below:$k$] { };
  \node[main] (f) [right=of k,label=above:$f$] { };
  \node[main] (z) [right=of f,label=below:$z$] {};
   \node[main, fill = white!100] (theta) [right=of  z,label=below:$\theta$] { };
   \node[main] (alpha) [right=of theta, label=below:$\alpha$] { };
   
  \node[main, fill = black!10] (y) [below=of f,label=left:$y$] { };
  \node[para, fill = black!100] (x) [right=of y,label=right:$x$] { };

  \path (alpha) edge [connect] (theta)
        (theta) edge [connect] (z)
		(z) edge [connect] (f)
		(k) edge [connect] (f)
		(f) edge [connect] (y)
		(x) edge [connect] (y)
		(beta) edge [connect] (lambda)
		(lambda) edge [connect] (k);
  \node[rectangle, inner sep=0mm, fit= (z),label=below right:D, xshift=-0.5mm,yshift=0.5mm] {};
  \node[rectangle, inner sep=4mm,draw=black!100, fit= (z)] {};
  \node[rectangle, inner sep=0mm, fit= (k),label=below right:L, xshift=0mm, yshift=0.5mm] {};
    \node[rectangle, inner sep=4mm,draw=black!100, fit= (k), yshift=0mm] {};
  \node[rectangle, inner sep=1mm, fit= (lambda) (k),label=above left:D, xshift=-5mm,yshift=-0.5mm] {};
  \node[rectangle, inner sep=5mm, draw=black!100, fit = (lambda) (k)] {};
\end{tikzpicture}
\caption{The graphical model for TileGP, a GP with additive and tile kernel partitioning structure. The parameter $\lambda$ controls the rate for the number of cuts $k$ of the tilings (inverse of the kernel bandwidth); the parameter $z$ controls the additive decomposition of the input feature space.}
\end{figure}

For the $j$-th hyperrectangle partition $\cx_j = [l_1^j, h_1^j]\times\cdots \times [l_D^j, h_D^j]$, we use a TileGP to model the function $f$ locally. We use the acronym ``TileGP'' to denote the Gaussian process model that uses additive kernels, with each component represented by tilings. We show the details of the generative model for TileGP in Alg.~\ref{alg:gen} and the graphical model in Fig.~\ref{fig:addgpgraph} with fixed hyper-parameters $\alpha, \beta_0, \beta_1$. The main difference to the additive GP model used in~\citep{wang2017batched} is that TileGP constructs a hierarchical model for the random features (and hence, the kernels), while \cite{wang2017batched} do not consider the kernel parameters to be part of the generative model. %
The random features are based on tile coding or Mondrian grids,  
with the number of cuts generated by $D$ Poisson processes on $[l_d^j,h_d^j]$ for each dimension $d=1,\cdots,D$. %
On the $i$-th layer of the tilings, tile coding samples the offset $\delta$ from a uniform distribution $U[0,\frac{h_d^j-l_d^j}{k_{di}}]$ and places the cuts uniformly starting at $\delta + l_d^j$. The Mondrian grid samples $k_{di}$ cut locations uniformly randomly from $[l_d^j,h_d^j]$. 
Because of the data partition, we always have more features than observations, which can alleviate the variance starvation problem described in Section~\ref{sec:var}.

We can use Gibbs sampling to efficiently learn the cut parameter $k$ and decomposition parameter $z$ by marginalizing out $\lambda$ and $\theta$. Notice that both $k$ and $z$ take discrete values; hence, unlike other continuous GP parameterizations, we only need to sample discrete variables for Gibbs sampling.
\begin{algorithm}[H]
  \caption{Generative model for TileGP}\label{alg:gen}
  \begin{small}
  \begin{algorithmic}[1]
\State Draw mixing proportions $\theta\sim\textsc{Dir}(\alpha)$
\For{$d = 1,\cdots, D$}
\State Draw additive decomposition $z_d\sim\textsc{Multi}(\theta)$
\State Draw Poisson rate parameter $\lambda_d\sim\textsc{Gamma}(\beta_0,\beta_1)$
\For{$i = 1, \cdots, L$}
\State \hspace{-1pt}Draw number of cuts $k_{di}\sim \textsc{Poisson}(\lambda_d(h_d^j-l_d^j))$ 
\State \hspace{-8pt}$\begin{cases} \text{Draw offset }\delta\sim U[0,\frac{h_d^j-l_d^j}{k_{di}}] & \text{Tile Coding}\\
\text{Draw cut locations } \vb\sim U[l_d^j,h_d^j] & \text{Mondrian Grids}
\end{cases}$
\EndFor
\EndFor
\State Construct the feature projection $\vpp$ and the kernel $\kk = \vpp\T \vpp$ from $z$ and sampled tiles
 \State Draw function $f\sim \cg\cp(0,\kk)$
\State Given input $\vx$, draw function value $y\sim \mathcal N(f(\vx), \sigma)$
\end{algorithmic}
\end{small}
\end{algorithm}
Given the observations $\cd_{t-1}$ in the $j$-th hyperrectangle partition, the posterior distribution of the (local) parameters $\lambda, k, z, \theta$ is 
\begin{align}
&p(\lambda,k,z,\theta\mid \cd_{t-1};\alpha,\beta) \nonumber \\
&\;\;\; \propto p(\cd_{t-1}\mid z,k) p(z\mid \theta)p(k\mid\lambda)p(\theta;\alpha)p(\lambda;\beta).\nonumber
\end{align}
Marginalizing over the Poisson rate parameter $\lambda$ and the mixing proportion $\theta$ gives 
\begin{align}
& p(k,z\mid \cd_{t-1};\alpha,\beta) \nonumber\\
& \propto p(\cd_{t-1} | z,k) \int p(z |\theta) p(\theta;\alpha)\dif \theta \int p(k| \lambda)p(\lambda;\beta)\dif\lambda \nonumber\\
&\propto p(\cd_{t-1}\mid z,k) \prod_m\frac{\Gamma(|A_m| + \alpha_m)}{\Gamma( \alpha_m)} \nonumber \\
& \;\;\;\;\;\; \times \prod_d \frac{\Gamma(\beta_1+|k_d|)}{(\prod_{i=1}^L k_{di}!)(\beta_0+L)^{\beta_1+|k_d|}} \nonumber
\end{align}
where $|k_d|=\sum_{i=1}^L k_{di}$. 
Hence, we only need to sample $k$ and $z$ when learning the hyperparameters of the TileGP kernel. For each dimension $d$, we sample the group assignment $z_d$ according to
\begin{align}
&p(z_d = m\mid \cd_{t-1}, k, z_{\neg d};\alpha) \propto p(\mathcal D_{t-1} \mid z,k) p(z_d\mid z_{\neg d}) \nonumber\\
& \;\;\; \propto p(\mathcal D_{t-1} \mid z,k) (|A_m| + \alpha_m).
\end{align}
We sample the number of cuts $k_{di}$ for each dimension $d$ and each layer $i$ from the posterior %
\begin{align}
& p(k_{di} \mid \cd_{t-1}, k_{\neg di}, z;\beta) \propto  p(\mathcal D_{t-1}\mid z,k) p(k_{di}\mid k_{\neg di})\nonumber\\
& \;\;\;\propto \frac{p(\mathcal D_n \mid z,k)\Gamma(\beta_{1}+|k_d|)}{(\beta_0+L)^{k_{di}}k_{di}!} .
\end{align}
If distributed computing is available, each hyperrectangle partition of the input space is assigned a worker to manage all the computations within this partition. On each worker, we use the above Gibbs sampling method to learn the additive structure and kernel bandwidth jointly. Conditioned on the observations associated with the partition on the worker, we use the learned posterior TileGP to select the most promising input point in this partition, and eventually send this candidate input point back to the main process together with the learned decomposition parameter $z$ and the cut parameter $k$. In the next section, we introduce the acquisition function we used in each worker and how to filter the recommended candidates from all the partitions.
\subsection{Acquisition functions and filtering}
\label{ssec:acfun}
In this paper, we mainly focus on parameter search problems where the objective function is designed by an expert and the global optimum or an upper bound on the function is known. While any BO acquisition functions can be used within the EBO framework, we use an acquisition function from~\citep{wang2017maxvalue} to exploit the knowledge of the upper bound. Let $f^*$ be such an upper bound, i.e., $\forall x\in \cx, f^*\geq f(x)$. Given the observations $\mathcal D^j_{t-1}$ associated with the $j$-th partition of the input space, we minimize the acquisition function $\eta^j_{t-1}(x) = \frac{f^* - \mu^j_{t-1}(x)}{\sigma^j_{t-1}(x)}$. Since the kernel is additive, we can optimize $\eta^j_{t-1}(\cdot)$ separately for each additive component. Namely, for the $m$-th component of the additive structure, we optimize $\eta^j_{t-1}(\cdot)$ only on the active dimensions $A_m$. This resembles a block coordinate descent, and greatly facilitates the optimization of the acquisition function. %

\paragraph{Filtering.} Once we have a proposed uery point from each partition, we select $B$ of them according to the scoring function $\xi(X) = \log \det K_X - \sum_{b=1}^B\eta(\vx_b)$ where $X=\{\vx_b\}_{b=1}^B$. We use the log determinant term to force diversity and $\eta$ to maintain quality. %
We maximize this function greedily. %
In some cases, the number of partitions $J$ can be smaller than the batch size $B$. In this case, one may either use just $J$ candidates, or use batch BO on each partition. We use the latter, and discuss details in the appendix.
\hide{
\subsection{Filtering selections and budget allocation}
In the EBO algorithm, we first use the batch workers to learn the local GPs and recommend potential good points from the local information. Then we aggregate the information of all the workers, and use a filter to select the points to evaluate from the set of points recommended by all the workers based on the aggregated information on the function. 

There are two important details here: (1) how many points to recommend from each local worker (budget allocation); and (2) how to filter the selections. Usually in the beginning of the iterations, we do not have a lot of leaves (since we stop partitioning a leaf once it reaches the minimum leaf size). Hence, it is very likely that the number of leaves $L$ is smaller than the size of the batch. Hence we need to allocate the budget of recommendations from each worker properly and use batch BO for each leaf. In our current version of EBO, we did the budget allocation using a heuristic, where we would like to generate at least $2B$ number of recommendations from all the workers, and each worker get the budget proportional to a score, the sum of the leaf volume (volume of the domain of the leaf) and the best function value of the leaf. For batch Bayesian optimization, we again uses a heuristic where  points achieving the top n acquisition function values is always included and the other ones come from random points selected in that leaf. 

Once we have all the recommended points, we select $B$ number of them by maximizing another acquisition function whose domain is a batch of points: $\xi(X) = \log \det K_X + \sum_{b=1}^B\eta(\vx_b)$ where $X=\{\vx_b\}_{b=1}^B$. We use the log determinant term to force diversity and $\eta$ to maintain quality. By the sub-modularity of the function $\xi$, we can optimize it greedily. Namely, we maximize $\xi(X\cup \{\vx\}) - \xi(X) = \log(\kk_\vx - \kk_{X\vx}\T K_X^{-1}\kk_{X\vx}) + \eta(\vx)$ for $B$ times. %
}
\begin{figure*}[t]
  \centering
\includegraphics[width=0.95\textwidth]{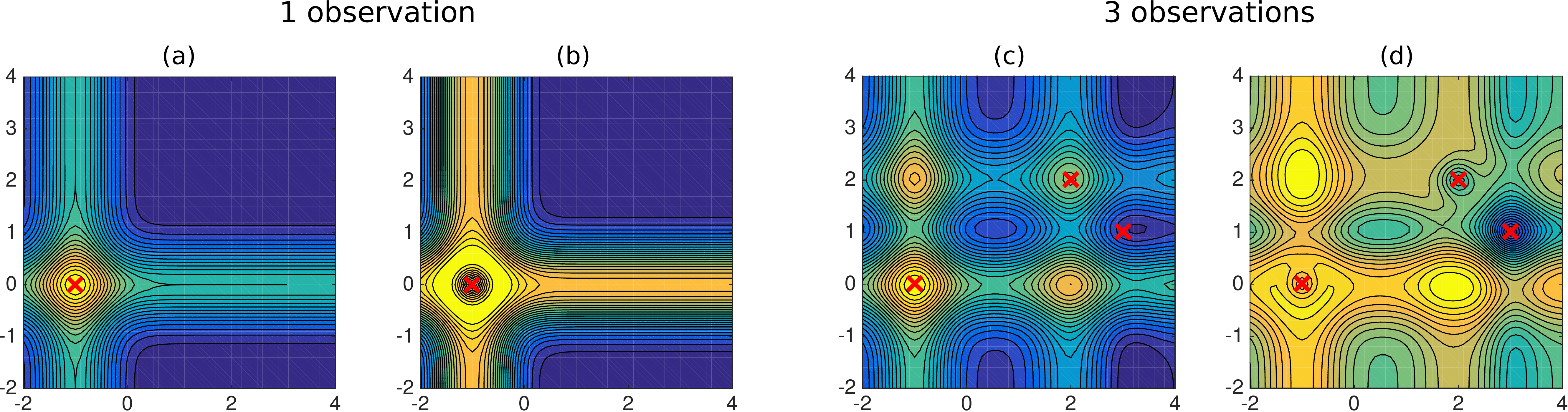} 
  \caption{Posterior mean function (a, c) and GP-UCB acquisition function (b, d) for an additive GP in 2D. The maxima of the posterior mean and acquisition function are at the points resulting from an exchange of coordinates between ``good'' observed points (-1,0) and (2,2).}
  \label{fig:cross}
\end{figure*}
\subsection{Efficient data likelihood computation and parameter synchronization}
For the random features, we use tile coding due to its sparsity and efficiency. Since non-zero features can be found and computed by binning, the computational cost for encoding a data point scales linearly with dimensions and number of layers. The resulting representation is sparse and convenient to use. %
Additionally, the number of non-zero features is quite small, which allows us to efficiently compute a sparse Cholesky decomposition of the inner product (Gram matrix) or the outer product of the data. This allows us to efficiently compute the data likelihoods. %

In each iteration $t$, after the batch workers return the learned decomposition indicator $z^b$ and the number of tiles $k^b, b\in[B]$, we synchronize these two parameters (line~\ref{sync} of Alg.~\ref{alg:ebo}). For the number of tiles $k$, we set $k_d$ to be the rounded mean of $\{k_d^b\}_{b=1}^B$ for each dimension $d\in[D]$. For the decomposition indicator, we use correlation clustering to cluster the input dimensions.

\subsection{Relations to Mondrian kernels, random binning and additive Laplace kernels}

Our model described in Section~\ref{ssec:gibbs} can use tile coding and Mondrian grids to construct the kernel. Tile coding and Mondrian grids are also closely related to Mondrian Features and Random Binning: All of the four kinds of random features attempt to find a sparse random feature representation for the raw input $\vx$ based on the partition of the space with the help of layers. We illustrate the differences between one layer of the features constructed by tile coding, Mondrian grids, Mondrian features and random binning in the appendix. Mondrian grids, Mondrian features and random binning all converge to the Laplace kernel as the number of layers $L$ goes to infinity. The tile coding kernel, however, does not approximate a Laplace kernel. Our model with Mondrian grids approximates an additive Laplace kernel:
\begin{lem}
Let the random variable $k_{di}\sim \textsc{Poisson}(\lambda_dR)$ be the number of cuts in the Mondrian grids of TileGP for dimension $d\in [D]$ and layer $i\in[L]$.  The TileGP kernel $\kk_L$ satisfies $\underset{L \to \infty}{\lim}\kk_L(\vx,\vx')= \frac{1}{M}\sum_{m=1}^Me^{\lambda_dR|\vx^{A_m}-\vx'^{A_m}|}$, where $\{A_m\}_{m=1}^M$ is the additive decomposition.  %
\end{lem}
We prove the lemma in the appendix. 
\cite{balog2016mondrian} showed that in practice, the Mondrian kernel constructed from Mondrian features may perform slightly better than random binning in certain cases. Although it would be possible to use a Mondrian partition for each layer of tile coding, we only consider uniform, grid based binning with random offests because this allows the non-zero features to be computed more efficiently ($O(1)$ instead of $O(\log k)$). Note that as more dimensions are discretized in this manner, the number of features grows exponentially. However, the number of non-zero entries can be independently controlled, allowing to create sparse representations that remain computationally tractable.

%% file: connection.tex
\subsection{Connections to evolutionary algorithms}

Next, we make some observations that connect our randomized ensemble BO to ideas for global optimization heuristics that have successfully been used in other communities. In particular, these connections offer an explanation from a BO perspective and may aid further theoretical analysis.

\emph{Evolutionary algorithms} \citep{back1996evolutionary} 
maintain an ensemble of ``good'' candidate solutions (called chromosomes) and, from those, generate new query points via a number of operations. These methods too, implicitly, need to balance exploration with local search in areas known to have high function values. Hence, there are local operations (mutations) for generating new points, such as random perturbations or local descent methods, and global operations. While it is relatively straightforward to draw connections between those local operations and optimization methods used in machine learning, we here focus on global exploration.

A popular global operation is \emph{crossover}: given two ``good'' points $x,y \in \mathbb{R}^D$, this operation outputs a new point $z$ whose coordinates are a combination of the coordinates of $x$ and $y$, i.e., $z_i \in \{x_i, y_i\}$ for all $i\in[D]$. In fact, this operation is analogous to BO with a (randomized) additive kernel: 
the crossover strategy implicity corresponds to the assumption that high function values can be achieved by combining coordinates from points with high function values. For comparison, consider an additive kernel $\kappa(x,x') = \sum_{m=1}^M \kappa^{(m)}(x^{A_m}, x'^{A_m})$ and $f(x) = \sum_{m=1}^Mf^m(x^{A_m})$. Since each sub-kernel $\kappa^{(m)}$ is ``blind'' to the dimensions in the complement of $A_m$, any point $x'$ that is close to an observed high-value point $x$ in the dimensions $A_m$ will receive a high value $f^m(x)$, independent of the other dimensions, and, as a result, looks like a ``good'' candidate.

We illustrate this reasoning with a 2D toy example. Figure~\ref{fig:cross} shows the posterior mean prediction and GP-UCB criterion $\hat{f}(x) + 0.1\sigma(x)$ for an additive kernel with $A_1 = \{1\}$, $A_2 = \{2\}$ and $\kappa^{m}(x_m,y_m) = \exp(-2(x_m-y_m)^2)$. High values of the observed points generalize along the dimensions ``ignored'' by the sub-kernels. After two good observations $(-1,0)$ and $(2,2)$, the ``crossover'' points $(-1,2)$ and $(2,0)$ are local maxima of GP-UCB and the posterior mean. 

In real data, we do not know the best fitting underlying grouping structure of the coordinates. Hence, crossover does a random search over such partitions by performing random coordinate combinations, whereas our adaptive BO approach maintains a posterior distribution over partitions that adapts to the data.

%% file: exp.tex
\section{Experiments}
We empirically verify the scalability of EBO and its effectiveness of using random adaptive Mondrian partitions, and finally evaluate EBO on two real-world problems.\footnote{Our code is publicly available at \url{https://github.com/zi-w/Ensemble-Bayesian-Optimization}.} %

\subsection{Scalability of EBO}
\label{ssec:exp:scalability}
\begin{figure}
 \centering
  \includegraphics[width=0.95\columnwidth]{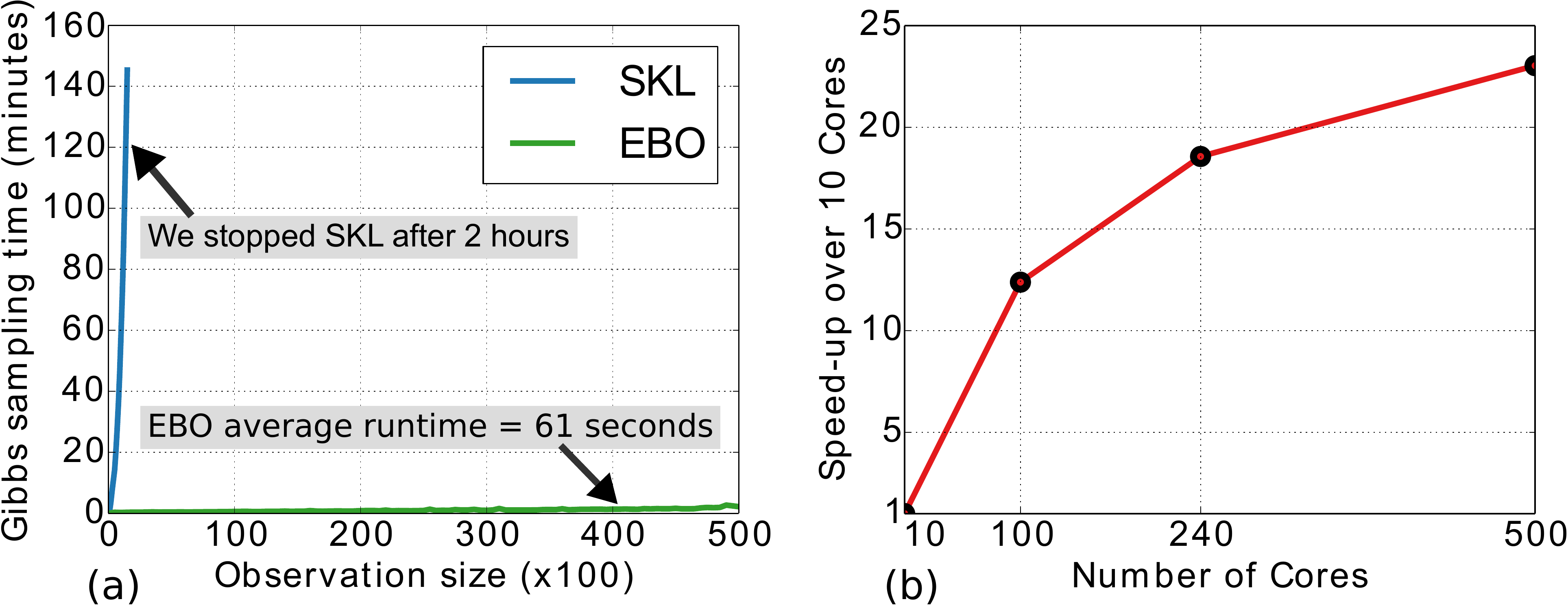}
 \caption{(a) Timing for the Gibbs sampler of EBO and SKL. EBO is significantly faster than SKL when the observation size $N$ is relatively large. (b) Speed-up of EBO with 100, 240, 500 cores over EBO with 10 cores on 30,000 observations. Running EBO with 240 cores is almost 20 times faster than with 10 cores. }
 \label{fig:timing}
 \end{figure}

 We compare EBO with a recent, state-of-the-art additive kernel learning algorithm, Structural Kernel Learning (SKL) \citep{wang2017batched}. EBO can make use of parallel resources both for Gibbs sampling and BO query selections, while SKL can only parallelize query selections but not sampling. Because the kernel learning part is the computationally dominating factor of large scale BO, we compare the time each method needs to run 10 iterations of Gibbs sampling with 100 to 50000 observations in 20 dimensions. We show the timing results for the Gibbs samplers in Fig.~\ref{fig:timing}(a), where EBO uses 240 cores via the Batch Service of Microsoft Azure. %
 Due to a time limit we imposed, we did not finish SKL for more than 1500 observations. EBO runs more than 390 times faster than SKL when the observation size is 1500. Comparing the quality of learned parameter $z$ for the additive structure, SKL has a Rand Index of $96.3\%$ and EBO has a Rand Index of $96.8\%$, which are similar. %
 In Fig.~\ref{fig:timing}(b), we show speed-ups for different number of cores. EBO with 500 cores is not significantly faster than with 240 cores because EBO runs synchronized parallelization, whose runtime is decided by the slowest core. It is often the case that most of the cores have finished while the program is waiting for the slowest 1 or 2 cores to finish.

\subsection{Effectiveness of EBO}
 \begin{figure}[h]
 \centering
  \includegraphics[width=0.65\columnwidth]{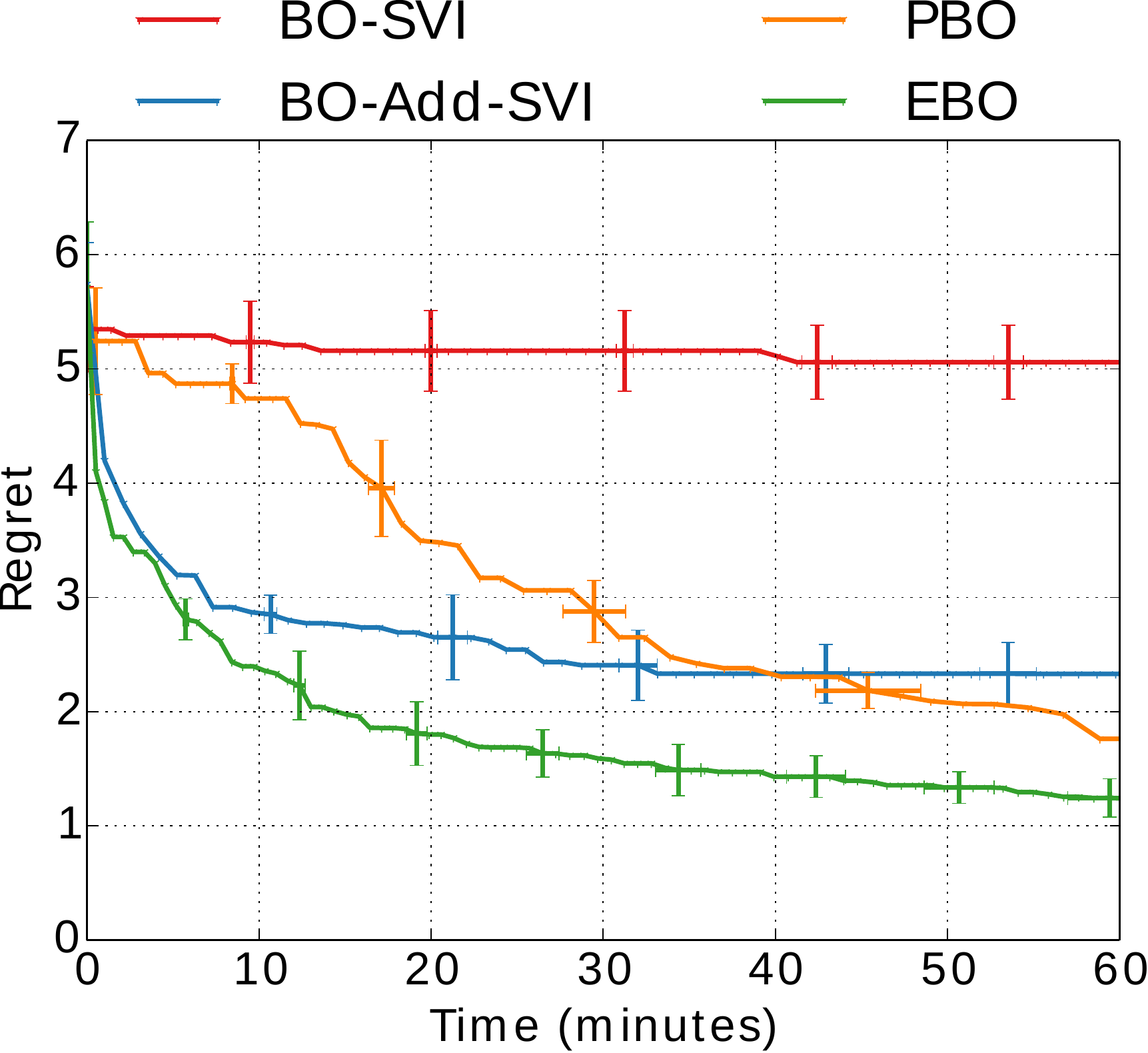}
 \caption{Averaged results of the regret of BO-SVI, BO-Add-SVI, PBO and EBO on 4 different functions drawn from a 50D GP with an additive Laplace kernel. BO-SVI has the highest regret for all functions. Using an additive GP within SVI (BO-Add-SVI) significantly improves over the full kernel. In general, EBO finds a good point much faster than the other methods.}
 \label{fig:ensemble}
 \end{figure}

\paragraph{Optimizing synthetic functions} We verify the effectiveness of using ensemble models for BO on 4 functions randomly sampled from a 50-dimensional GP with an additive Laplace kernel. The hyperparameter of the Laplace kernel is known. In each iteration, each algorithm evaluates a batch of parameters of size $B$ in parallel. We denote $\tilde r_{t} = \max_{x\in \cx}f(x) - \max_{ b\in[B]}f(x_{t,b})$ as the immediate regret obtained by the batch at iteration $t$, and $r_{T} = \min_{t\leq T}  \tilde r_{t}$ as  the regret, which captures the minimum gap between the best point found and the global optimum of the black-box function $f$.

We compare BO using SVI~\citep{hensman2013gaussian} (BO-SVI), BO using SVI with an additive GP (BO-Add-SVI) and a distributed version of BO with a fixed partition (PBO) against EBO with a randomly sampled partition in each iteration. PBO has the same 1000 Mondrian partitions in all the iterations while EBO can have at most 1000 Mondrian partitions. BO-SVI uses a Laplace isotropic kernel without any additive structure, while BO-Add-SVI, PBO, EBO all use the known prior. More detailed experimental settings can be found in the appendix. %
Our experimental results in Fig.~\ref{fig:ensemble} shows that EBO is able to find a good point much faster than BO-SVI and BO-Add-SVI; and, randomization and the ensemble of partitions matters: EBO is much better than PBO.

\paragraph{Optimizing control parameters for robot pushing} We follow~\cite{wang2017batched} and test our approach, EBO, on a 14 dimensional control parameter tuning problem for robot pushing. We compare EBO, BO-SVI, BO-Add-SVI and CEM~\cite{szita2006learning} with the same $10^4$ random observations and repeat each experiment $10$ times.  We run all the methods for 200 iterations, where each iteration has a batch size of $100$. We plot the median of the best rewards achieved by CEM and EBO at each iteration in Fig.~\ref{fig:robotpush}. More details on the experimental setups and the reward function can be found in the appendix. 
\hide{
We implemented the simulation of pushing two objects with two robot hands in the Box2D physics engine~\cite{box2d}. The 14 parameters specifies the location and rotation of the robot hands, pushing speed, moving direction and pushing time. The lower limit of these parameters is $[-5, -5, -10, -10, 2, 0, -5, -5, -10, -10, 2, 0, -5, -5]$ and the upper limit is $[5, 5, 10, 10, 30, 2\pi, 5, 5, 10, 10, 30, 2\pi, 5, 5]$. Let the initial positions of the objects be $s_{i0}, s_{i1}$ and the ending positions be $s_{e0}, s_{e1}$. We use $s_{g0}$ and $s_{g1}$ to denote the goal locations for the two objects. The reward is defined to be $r = \|s_{g0}-s_{i0} \| +\|s_{g1}-s_{i1} \| - \|s_{g0}-s_{e0} \| -\|s_{g1}-s_{e1} \|$, namely, the progress made towards pushing the objects to the goal. 

We compare EBO, BO-SVI, BO-Add-SVI and CEM~\cite{szita2006learning} with the same $10^4$ random observations and repeat each experiment $10$ times. All the methods choose a batch of $100$ parameters to evaluate at each iteration. CEM uses the top $30\%$ of the $10^4$ initial observations to fit its initial Gaussian distribution. At the end of each iteration in CEM, $30\%$ of the new observations with top values were used to fit the new distribution. For all the BO based methods, we use the maximum value of the reward function in the acquisition function. The standard deviation of the observation noise in the GP models is set to be $0.1$. We set EBO to have Modrian partitions with fewer than 150 data points and constrain EBO to have no more than 200 Mondrian partitions. In EBO, we set the hyper parameters $\alpha=1.0$,  $\beta=[5.0,5.0]$, and the Mondrian observation offset $\epsilon=0.05$. In BO-SVI, we used $100$ batchsize in SVI, $200$ inducing points and $500$ iterations to optimize the data likelihood with $0.1$ step rate and $0.9$ momentum. BO-Add-SVI used the same parameters as BO-SVI, except that BO-Add-SVI uses 3 outer loops to randomly select the decomposition parameter $z$ and in each loop, it uses an inner loop of $50$ iterations to maximize the data likelihood over the kernel parameters. The batch BO strategy used in BO-SVI and BO-Add-SVI is identical to the one used in each Mondrian partition of EBO. 

 We run all the methods for 200 iterations, where each iteration has a batch size of $100$. In total, each method obtains $2\times 10^4$ data points in addition to the $10^4$ initializations. We repeat the experiments 10 times and plot the median of the best rewards achieved by CEM and EBO at each iteration in Fig.~\ref{fig:robotpush}. 
 }
 Overall CEM and EBO performed comparably and much better than the sparse GP methods (BO-SVI and BO-Add-SVI). We noticed that among all the experiments, CEM achieved a maximum reward of $10.19$ while EBO achieved $9.50$. However, EBO behaved slightly better and more stable than CEM as reflected by the standard deviation on the rewards. 
 \begin{figure}[h!]
 \centering
  \includegraphics[width=0.65\columnwidth]{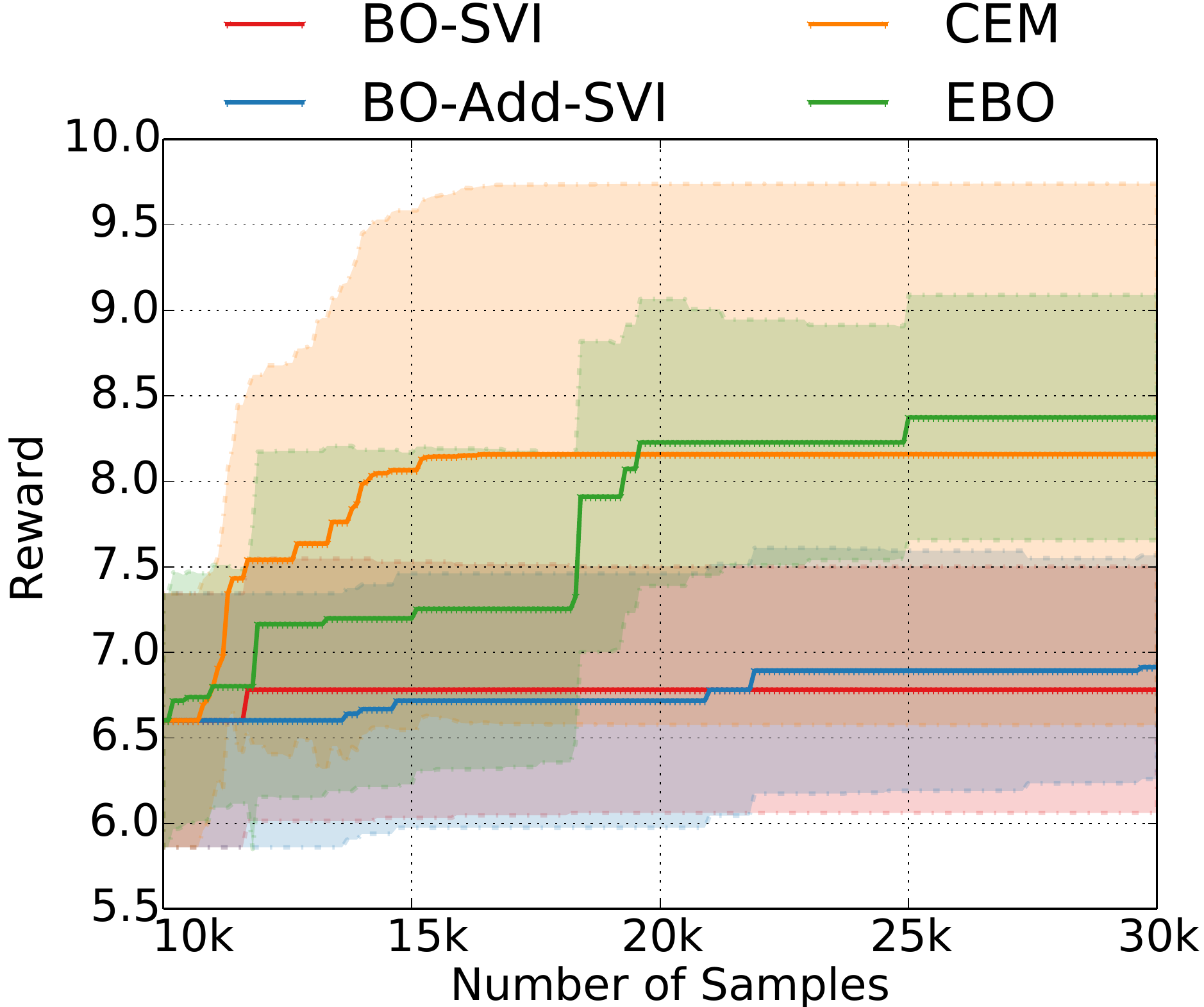}
 \caption{Comparing BO-SVI, BO-Add-SVI, CEM and EBO on a control parameter tuning task with 14 parameters. }
 \label{fig:robotpush}
 \end{figure}

%% file: rover.tex
\paragraph{Optimizing rover trajectories}

\begin{figure}[h!]
 \centering
  \includegraphics[width=0.65\columnwidth]{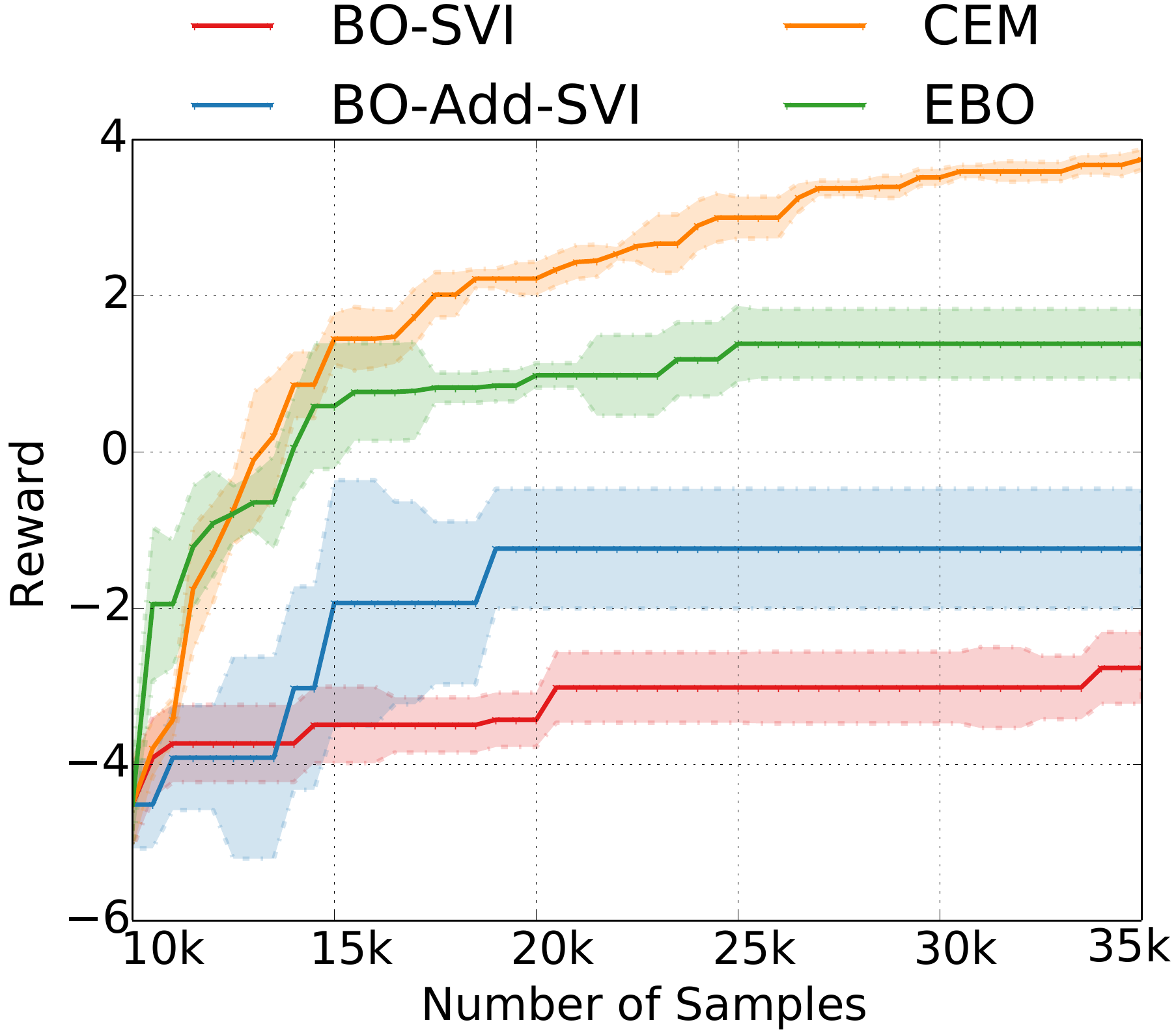}
 \caption{Comparing BO-SVI, BO-Add-SVI, CEM and EBO on a $60$ dimensional trajectory optimization task.}
 \label{fig:rover}
 \end{figure}

To further explore the performance of our method, we consider a trajectory optimization task in 2D, meant to emulate a rover navigation task. We describe a problem instance by defining a start position $s$ and a goal position $g$ as well as a cost function over the state space. Trajectories are described by a set of points on which a BSpline is to be fitted. By integrating the cost function over a given trajectory, we can compute the trajectory cost $c(\vx)$ of a given trajectory solution $\vx\in [0,1]^{60}$.  We define the reward of this problem to be $f(\vx) = c(\vx) + \lambda(\|\vx_{0,1}-s\|_1 + \|\vx_{59,60}-g\|_1) + b$. This reward function is non smooth, discontinuous, and concave over the first two and last two dimensions of the input. These 4 dimensions represent the start and goal position of the trajectory. The results in Fig.~\ref{fig:rover} showed that CEM was able to achieve better results than the BO methods on these functions, while EBO was still much better than the BO alternatives using SVI. 
More details can be found in the appendix. 

%% file: conclu.tex
\section{Conclusion}
Many black box function optimization problems are intrinsically high-dimensional and may require a huge number of observations in order to be optimized well. In this paper, we propose a novel framework, ensemble Bayesian optimization, to tackle the problem of scaling Bayesian optimization to both large numbers of observations and high dimensions. To achieve this, we propose a new framework that jointly integrates randomized partitions at various levels: our method is a stochastic method over a randomized, adaptive ensemble of partitions of the input data space; for each part, we use an ensemble of TileGPs, a new GP model we propose based on tile coding and additive structure. We also developed an efficient Gibbs sampling approach to learn the latent variables. Moreover, our method automatically generates batch queries. We empirically demonstrate the effectiveness and scalability of our method on high dimensional parameter search tasks with tens of thousands of observation data. %

%% file: sup_method.tex
\section{An Illustration of EBO}

\begin{figure}[h]
  \includegraphics[width=0.8\columnwidth]{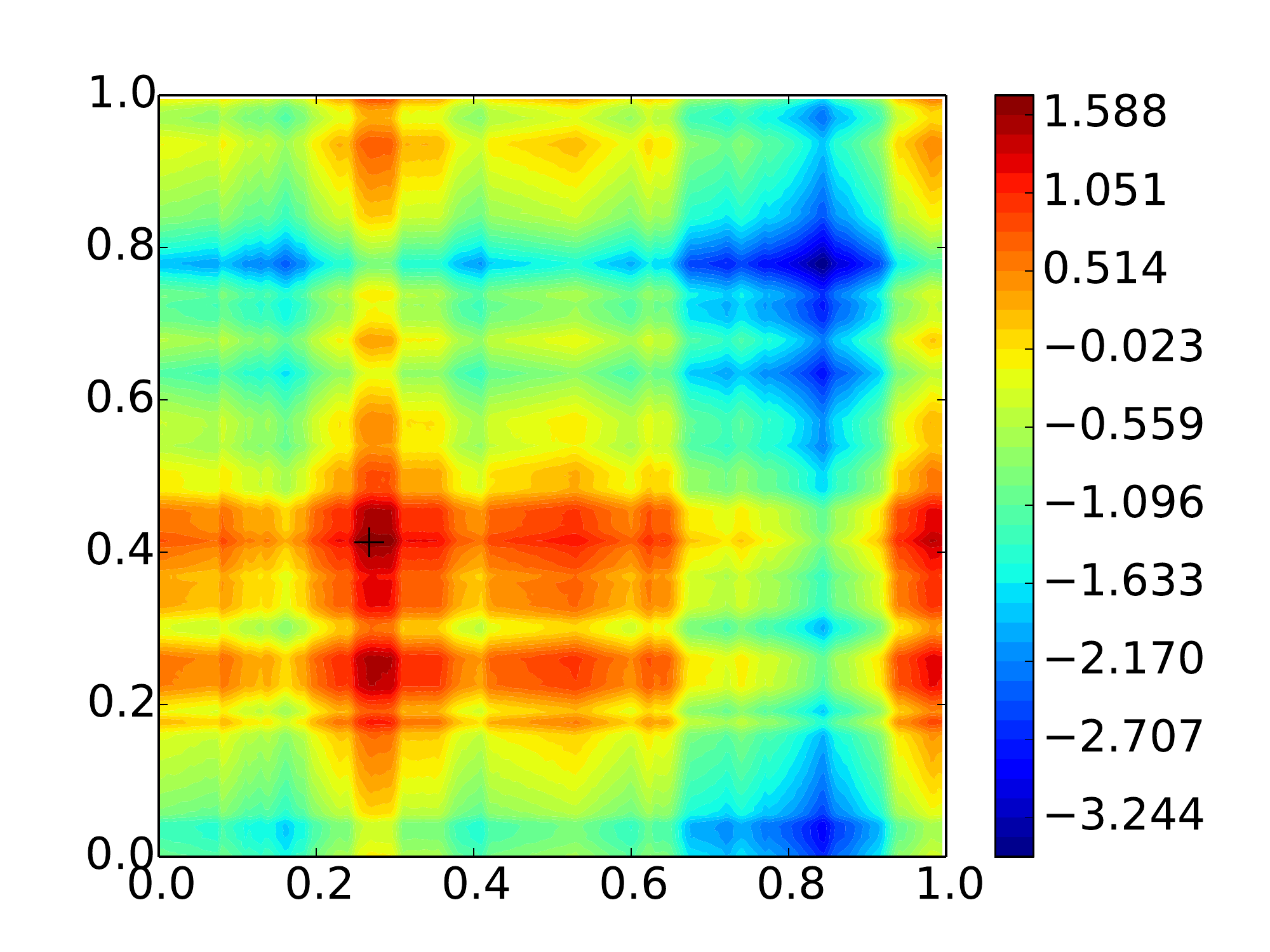}
 \caption{The 2D additive function we optimized in Fig.~\ref{fig:example}. The global maximum is marked with ``+''.}
 \label{fig:func}
\end{figure}

We give an illustration of the proposed EBO algorithm on a 2D function shown in Fig.~\ref{fig:func}. This function is a sample from a 2D TileGP, where the decomposition parameter is $z=[0,1 ]$, the cut parameter is (inverse bandwidth) $k=[10,10]$, and the noise parameter is $\sigma=0.01$. 

\begin{figure*}[t]
 \centering
  \includegraphics[width=1\textwidth]{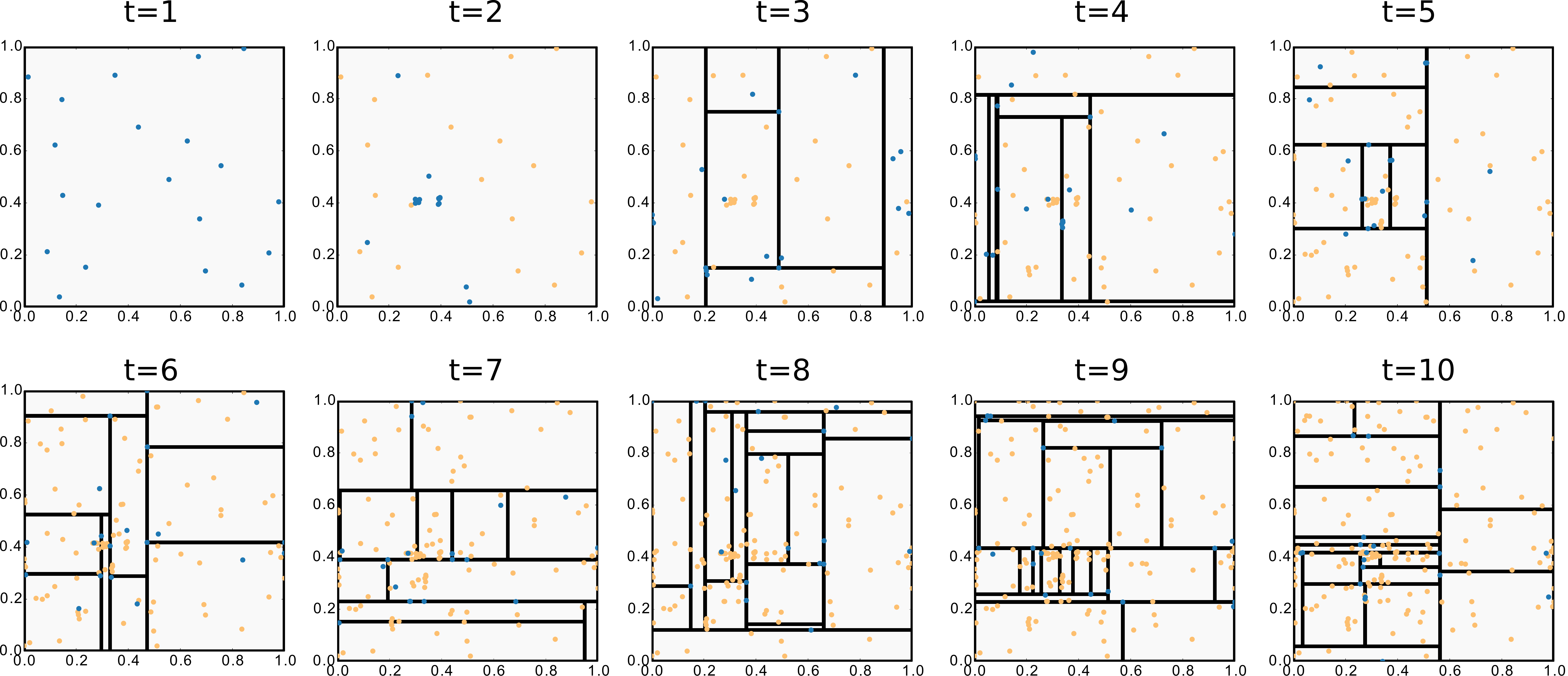}
 \caption{An example of 10 iterations of EBO on a 2D toy example plotted in Fig.~\ref{fig:func}. The selections in each iteration are blue and the existing observations orange. EBO quickly locates the region of the global optimum while still allocating budget to explore regions that appear promising (e.g. around the local optimum $(1.0,0.4)$). }
 \label{fig:example}
\end{figure*}

The global maximum of this function is at $(0.27, 0.41)$. In this example, EBO is configured to have at least 20 data points on each partition, at most 50 Mondrian partitions, and 100 layers of tiles to approximate the Laplace kernel. We run EBO for 10 iterations with 20 queries each batch. The results are shown in Fig.~\ref{fig:example}. In the first iteration, EBO has no information about the function; hence it spreads the 10 queries (blue dots) ``evenly'' in the input domain to collect information. In the 2nd iteration, based on the evaluations on the selected points (yellow dots), EBO chooses to query batch points (blue dots) that have high acquisition values, which appear to be around the global optimum and some other high valued regions. As the number of evaluations exceeds 20, the minimum number of data points on each partition, EBO partitions the input space with a Mondrian process in the following iterations. Notice that each iteration draws a different partition (shown as the black lines) from the Mondrian process so that the results will not ``over-fit'' to one partition setting and the computation can remain efficient. In each partition, EBO runs the Gibbs sampling inference algorithm to fit a local TileGP and uses batched BO select a few candidates. Then EBO uses a filter to decide the final batch of candidate queries (blue dots) among all the recommended ones from each partition as described in Sec.~\ref{sec:filter}.

\section{Partitioning the input space via a Mondrian process}
Alg.~\ref{alg:mondrian} shows the full `Mondrian partitioning'' algorithm, i.e., the input space partitioning strategy mentioned in Section 3.1. 
\begin{algorithm}[H]
  \caption{Mondrian Partitioning }\label{alg:mondrian}
    \begin{small}
  \begin{algorithmic}[1]
    \Function{MondrianPartitioning\,}{$V, N_{p}, S$}
      \While{$|V| <  N_{p}$}
      \State  $p_j\gets length(v_j) \cdot max(0, |\cd^j| - S), \forall v_j \in V$
      \If{$p_j = 0, \forall j$}
      \State break
      \EndIf
      \State Sample $ v_j \sim \frac{p_j}{\sum_j p_j}, v_j \in V$
      \State Sample a dimension $ d \sim \frac{h_d^j - l_d^j}{\sum_d h_d^j - l_d^j}, d\in [D]$
      \State Sample cut location $u_d^j \sim U[l_d^j, h_d^j]$
      \State $v_{j(left)}\gets [l_1^j, h_1^j]\times\cdots \times [l_d^j, u_d^j]\times \cdots\times \times [l_D^j, h_D^j]$
      \State $v_{j(right)}\gets [l_1^j, h_1^j]\times\cdots \times [u_d^j, h_d^j]\times \cdots\times \times [l_D^j, h_D^j]$
      \State $V \gets V\cup\{v_{j(left)},v_{j(right)}\}\setminus v_j$
      \EndWhile
      \State\Return $V$
      \EndFunction
  \end{algorithmic}
        \end{small}
\end{algorithm}
In particular, we denote the maximum number of Mondrian partitions by $N_{p}$ (usually the worker pool size in the experiments) and the minimum number of data points in each partition to be $S$. The set of partitions computed by the Mondrian tree (a.k.a. the leaves of the tree), $V$, is initialized to be the function domain $V=\{[0,R]^D\}$, the root of the tree. For each $v_j\in V$ described by a hyperrectangle $[l_1^j, h_1^j]\times\cdots \times [l_D^j, h_D^j]$, the length of $v_j$ is computed to be $length(v_j) = \sum_{d=1}^D (h_d^j-l_d^j)$. The observations associated with $v_j$ is $\cd^j$. Here, for all  $(x,y)\in \cd^j$, we have $x\in [l_1^j-\epsilon, h_1^j+\epsilon]\times\cdots \times [l_D^j-\epsilon, h_D^j+\epsilon]$, where $\epsilon$ controls the how many neighboring data points to consider for the partition $v_j$. In our experiments, $\epsilon$ is set to be $0$. Alg.~\ref{alg:mondrian} is different from Algorithm 1 and 2 of~\cite{lakshminarayanan2016mondrian} in the stop criterion. \cite{lakshminarayanan2016mondrian} uses an exponential clock to count down the time of splitting the leaves of the tree, while we split the leaves until the number of Mondrian partitions reaches $N_{p}$ or there is no partition that have more than $S$ data points. We designed our stop criterion this way to balance the efficiency of EBO and the quality of selected points. Usually EBO is faster with larger number of partitions $N_{p}$ (i.e., more parallel computing resources) and the quality of the selections are better with larger size of observations on each partition ($S$).

\section{Budget allocation and batched BO}
\label{sec:filter}
In the EBO algorithm, we first use a batch of workers to learn the local GPs and recommend potential good candidate points from the local information. Then we aggregate the information of all the workers, and use a filter to select the points to evaluate from the set of points recommended by all the workers based on the aggregated information on the function. 

There are two important details we did not have space to discuss in the main paper: (1) how many points to recommend from each local worker (budget allocation); and (2) how to select a batch of points from the Mondrian partition on each worker. Usually in the beginning of the iterations, we do not have a lot of Mondrian partitions (since we stop splitting a partition once it reaches a minimum number of data points). Hence, it is very likely that the number of partitions $J$ is smaller than the size of the batch. Hence we need to allocate the budget of recommendations from each worker properly and use batched BO for each Mondrian partition. 

\paragraph{Budget allocation} In our current version of EBO, we did the budget allocation using a heuristic, where we would like to generate at least $2B$ recommendations from all the workers, and each worker gets the budget proportional to a score, the sum of the Mondrian partition volume (volume of the domain of the partition) and the best function value of the partition. 

\paragraph{Batched BO} For batched BO, we also use a heuristic where the  points achieving the top $n$ acquisition function values are always included and the other ones come from random points selected in that partition. For the optimization of the acquisition function over each block of dimensions, we sample 1000 points in the low dimensional space associated with the additive component and minimize the acquisition function via L-BFGS-B starting from the point that gives the best acquisition value. We add the optimized $\argmin$ to the 1000 points and sort them according to their acquisition values, and then select the top $n$ random ones, and combine with the sorted selections from other additive components. Other batched BO methods can also be used and can potentially improve upon our results.
\section{Relations to Mondrian Kernels and Random Binning}
TileGP can use Mondrian grids or (our version of) tile coding to achieve efficient parameter inference for the decomposition $z$ and the number of cuts $k$ (inverse of kernel bandwidth). Mondrian grids and tile coding are closely related to Mondrian kernels and random binning, but there are some subtle differences. 
 \begin{figure}[h!]
 \centering
  \includegraphics[width=\columnwidth]{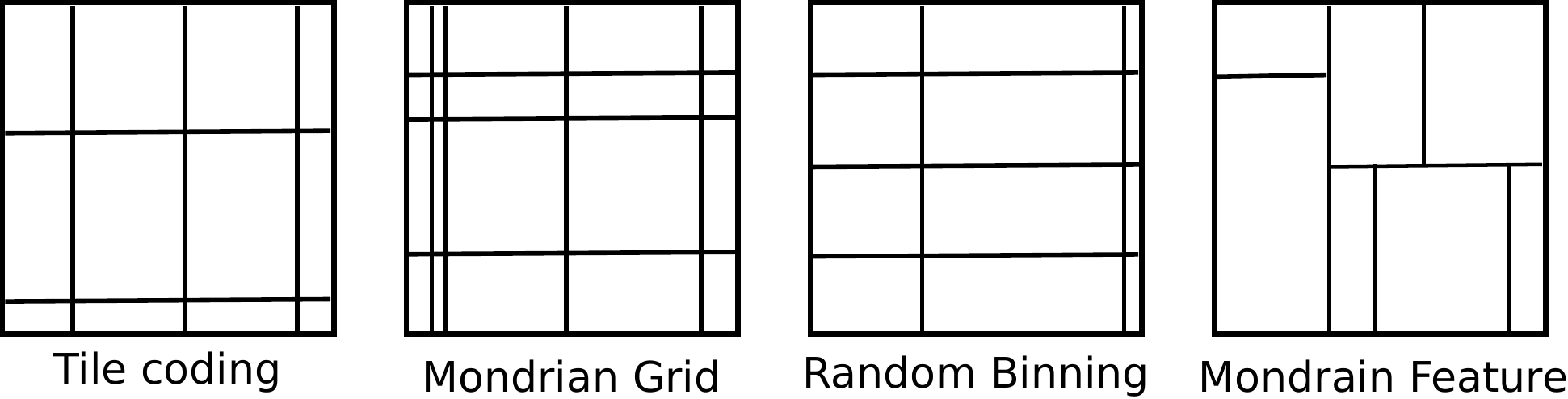}
 \caption{Illustrations of (our version of) tile coding, Mondrian Grid, random binning and Mondrian feature.}
 \label{fig:feature}
 \end{figure}
We illustrate the differences between one layer of the features constructed by tile coding, Mondrian grid, Mondrian feature and random binning in Fig.~\ref{fig:feature}. For each layer of (our version of) tile coding, we sample a positive integer $k$ (number of cuts) from a Poisson distribution parameterized by $\lambda R$, and then set the offset to be a constant uniformly randomly sampled from $[0, \frac{R}{k}]$. For each layer of the Mondrian grid, the number of cuts $k$ is sampled tile in coding, but instead of using an offset and uniform cuts, we put the cuts at locations independently uniformly randomly from $[0,R]$. Random binning does not sample $k$ cuts but samples the distance $\delta$ between neighboring cuts by drawing $\delta\sim \textsc{Gamma}(2,\lambda R)$. Then, it samples the offset from $[0,\delta]$ and finally places the cuts. All of the above-mentioned three types of random features can work individually for each dimension and then combine the cuts from all dimensions. The Mondrian feature (Mondrian forest features to be exact), contrast, partitions the space jointly for all dimensions. More details of Mondrian features can be found in~\cite{lakshminarayanan2016mondrian,balog2016mondrian}. For all of these four types of random features and for each layer of the total $L$ layers, the kernel is $\kk_L(\vx,\vx') = \frac{1}{L}\sum_{l=1}^L \chi_l(\vx,\vx')$ where 
\begin{align}
\chi_l(\vx, \vx') = \begin{cases}
1 & \text{$\vx$ and $\vx'$ are in the same cell on the layer $l$}\\
0 & \text{otherwise}
\end{cases}
\end{align}
For the case where the kernel has $M$ additive components, we simply use the tiling for each decomposition and normalize by $LM$ instead of $L$. More precisely, we have $\kk_L(\vx,\vx') =  \frac{1}{LM}\sum_{m=1}^M\sum_{l=1}^M \chi_l(\vx^{A_m},\vx'^{A_m})$.
 
 We next prove the lemma mentioned in Section 3.5.
\begin{customlemma}{3.1}
Let the random variable $k_{di}\sim \textsc{Poisson}(\lambda_dR)$ be the number of cuts in the Mondrian grids of TileGP for dimension $d\in [D]$ and layer $i\in[L]$.  The kernel of TileGP $\kk_L$ satisfies $\underset{L \to \infty}{\lim}\kk_L(\vx,\vx')= \frac{1}{M}\sum_{m=1}^Me^{\lambda_dR|\vx^{A_m}-\vx'^{A_m}|}$, where $\{A_m\}_{m=1}^M$ is the additive decomposition. %
\end{customlemma}
\begin{proof}
When constructing the Mondrian grid for each layer and each dimension, one can think of the process of getting another cut as a Poisson point process on the interval $[0,R]$, where the time  between two consecutive cuts is modeled as an exponential random variable. Similar to Proposition 1 in~\cite{balog2016mondrian}, we have $\underset{L\to\infty}{\lim}{\kk_L^{(m)} (\vx^{A_m},\vx'^{A_m})} = \Ex[\text{no cut between $\vx_d$ and $\vx'_d, \forall d\in A_m$}] = e^{-\lambda_d R|\vx^{A_m}-\vx'^{A_m}|}$. By the additivity of the kernel, we have $\underset{L \to \infty}{\lim}\kk_L(\vx,\vx')= \frac{1}{M}\sum_{m=1}^Me^{\lambda_dR|\vx^{A_m}-\vx'^{A_m}|}$.
\end{proof}

%% file: sup_exp.tex
\section{Experiments}
\begin{figure}[h!]
 \centering
  \includegraphics[width=\columnwidth]{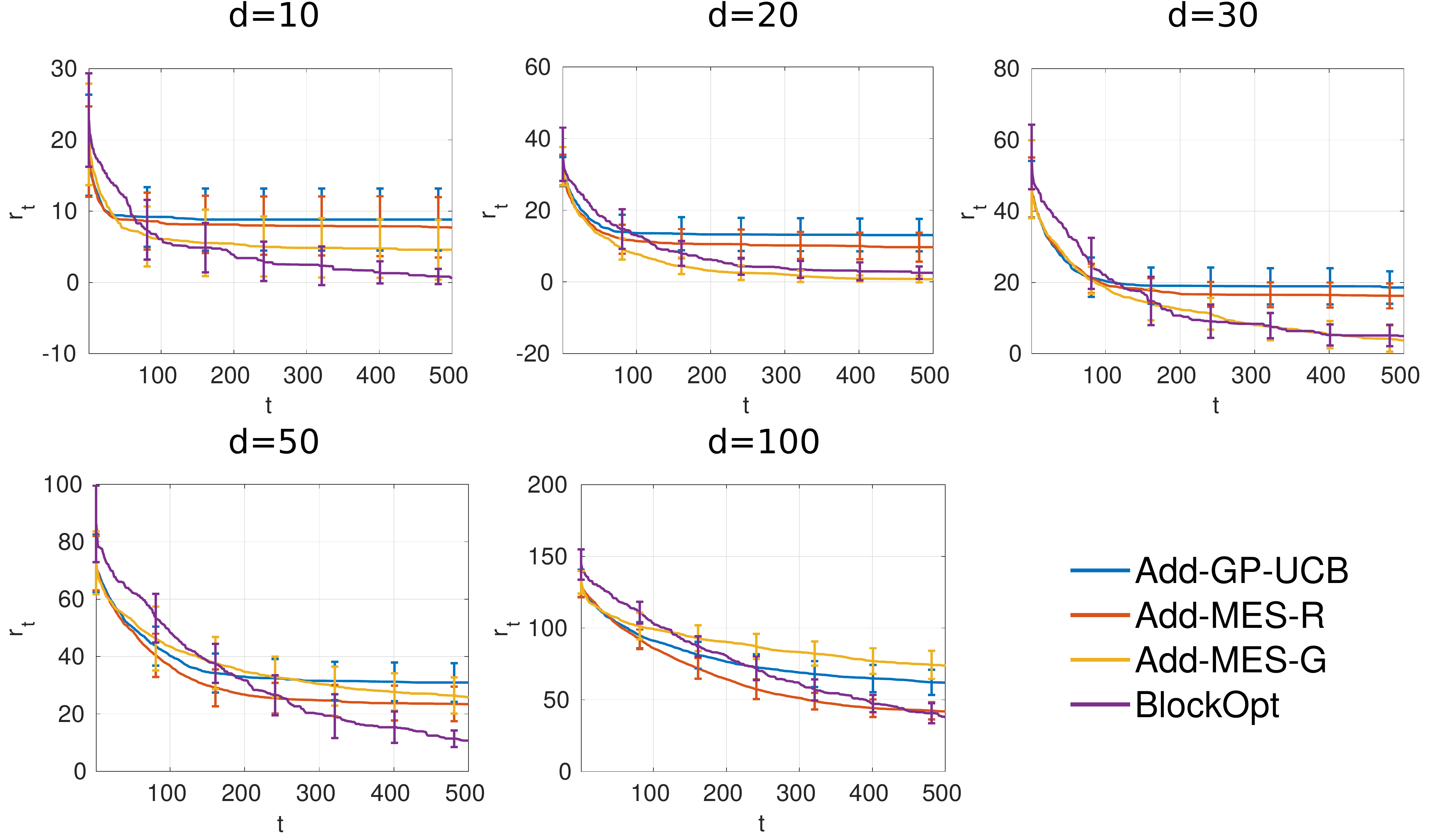}
 \caption{Comparing different acquisition functions for BO with an additive GP. Our strategy, BlockOpt, achieves comparable or better results than other methods.}
 \label{fig:high}
 \end{figure}

\paragraph{Verifying the acquisition function} As introduced in Section 3.3, we used a different acquisition function optimization technique from~\citep{kandasamy2015high,wang2017maxvalue}. In~\citep{kandasamy2015high,wang2017maxvalue}, the authors used the fact that each additive component is by itself a GP. Hence, they did posterior inference on each additive component and Bayesian optimization independently from other additive components. In this work, we use the full GP with the additive kernel to derive its acquisition function and optimize it with a block coordinate optimization procedure, where the blocks are selected according to the decomposition of the input dimensions.  One reason we did this instead of following~\citep{kandasamy2015high,wang2017maxvalue} is that we observed the over-estimation of variance for each additive component if inferred independently from others. We conjecture that this over-estimation could result in an invalid regret bound for Add-GP-UCB~\citep{kandasamy2015high}. Nevertheless, we found that using the block coordinate optimization for the acquisition function on the full GP is actually very helpful. In Figure.~\ref{fig:high}, we compare the acquisition function we described in Section 3.3 (denoted as BlockOpt) with Add-GP-UCB~\citep{kandasamy2015high}, Add-MES-R and Add-MES-G~\citep{wang2017maxvalue} on the same experiment described in the first experiment of Section 6.5 of~\citep{wang2017maxvalue}, averaging over 20 functions. Notice that we used the maximum value of the function as part of our acquisition function in our approach (BlockOpt). Add-GP-UCB, ADD-MES-R and ADD-MES-G cannot use this max-value information even if they have access to it, because then they don't have a strategy to deal with ``credit assignment'', which assigns the maximum value to each additive component. We found that BlockOpt is able to find a solution as well as or even better than the best of the three competing approaches. 

\paragraph{Scalability of EBO} For EBO, the maximum number of Mondrian partitions is set to be 1000 and the minimum number of data points in each Mondrian partition is 100. The function that we used to test was generated from a fully partitioned 20 dimensional GP with an additive Laplace kernel ($|A_m|=1,\forall m$).
\paragraph{Effectiveness of EBO} In this experiment, we sampled 4 functions from a 50-dimensional GP with additive kernel. Each component of the additive kernel is a Laplace kernel, whose lengthscale parameter is set to be $0.1$, variance scale to be $1$ and active dimensions are around 1 to 4. Namely, the kernel we used is $\kk(x,x') = \sum_{i=1}^M \kk^{(m)}(x^{A_m}, x'^{A_m})$ where $\kk^{(m)}(x^{A_m}, x'^{A_m}) = e^{\frac{|x^{A_m}- x'^{A_m}|}{0.1}}, \forall m$. The domain of the function is $[0,1]^{50}$. We implemented the BO-SVI and BO-Add-SVI using the same acquisition function and batch selection strategy as EBO but with SVI-GP~\citep{hensman2013gaussian} and SVI-GP with additive kernels instead of TileGPs. We used the SVI-GP implemented in~\cite{gpy2014} and defined the additive Laplace kernel according to the priors of the tested functions. For both BO-SVI and BO-Add-SVI, we used 100 batchsize, 200 inducing points and the parameters were optimized for 100 iterations. For EBO, we set the minimum size of data points on each Mondrian partition to be 100. We set the maximum number of Mondrian partitions to be 1000 for both EBO and PBO. The evaluations of the test functions are negligible, so the timing results in Figure 5 reflect the actual runtime of each method.

\paragraph{Optimizing control parameters for robot pushing} 
We implemented the simulation of pushing two objects with two robot hands in the Box2D physics engine~\cite{box2d}. The 14 parameters specifies the location and rotation of the robot hands, pushing speed, moving direction and pushing time. The lower limit of these parameters is $[-5, -5, -10, -10, 2, 0, -5, -5, -10, -10, 2, 0, -5, -5]$ and the upper limit is $[5, 5, 10, 10, 30, 2\pi, 5, 5, 10, 10, 30, 2\pi, 5, 5]$. Let the initial positions of the objects be $s_{i0}, s_{i1}$ and the ending positions be $s_{e0}, s_{e1}$. We use $s_{g0}$ and $s_{g1}$ to denote the goal locations for the two objects. The reward is defined to be $r = \|s_{g0}-s_{i0} \| +\|s_{g1}-s_{i1} \| - \|s_{g0}-s_{e0} \| -\|s_{g1}-s_{e1} \|$, namely, the progress made towards pushing the objects to the goal. 

We compare EBO, BO-SVI, BO-Add-SVI and CEM~\cite{szita2006learning} with the same $10^4$ random observations and repeat each experiment $10$ times. All the methods choose a batch of $100$ parameters to evaluate at each iteration. CEM uses the top $30\%$ of the $10^4$ initial observations to fit its initial Gaussian distribution. At the end of each iteration in CEM, $30\%$ of the new observations with top values were used to fit the new distribution. For all the BO based methods, we use the maximum value of the reward function in the acquisition function. The standard deviation of the observation noise in the GP models is set to be $0.1$. We set EBO to have Modrian partitions with fewer than 150 data points and constrain EBO to have no more than 200 Mondrian partitions. In EBO, we set the hyper parameters $\alpha=1.0$,  $\beta=[5.0,5.0]$, and the Mondrian observation offset $\epsilon=0.05$. In BO-SVI, we used $100$ batchsize in SVI, $200$ inducing points and $500$ iterations to optimize the data likelihood with $0.1$ step rate and $0.9$ momentum. BO-Add-SVI used the same parameters as BO-SVI, except that BO-Add-SVI uses 3 outer loops to randomly select the decomposition parameter $z$ and in each loop, it uses an inner loop of $50$ iterations to maximize the data likelihood over the kernel parameters. The batch BO strategy used in BO-SVI and BO-Add-SVI is identical to the one used in each Mondrian partition of EBO. 

 We run all the methods for 200 iterations, where each iteration has a batch size of $100$. In total, each method obtains $2\times 10^4$ data points in addition to the $10^4$ initializations.

 \paragraph{Optimizing rover trajectories}

\begin{figure}
  \centering
   \includegraphics[width=\columnwidth]{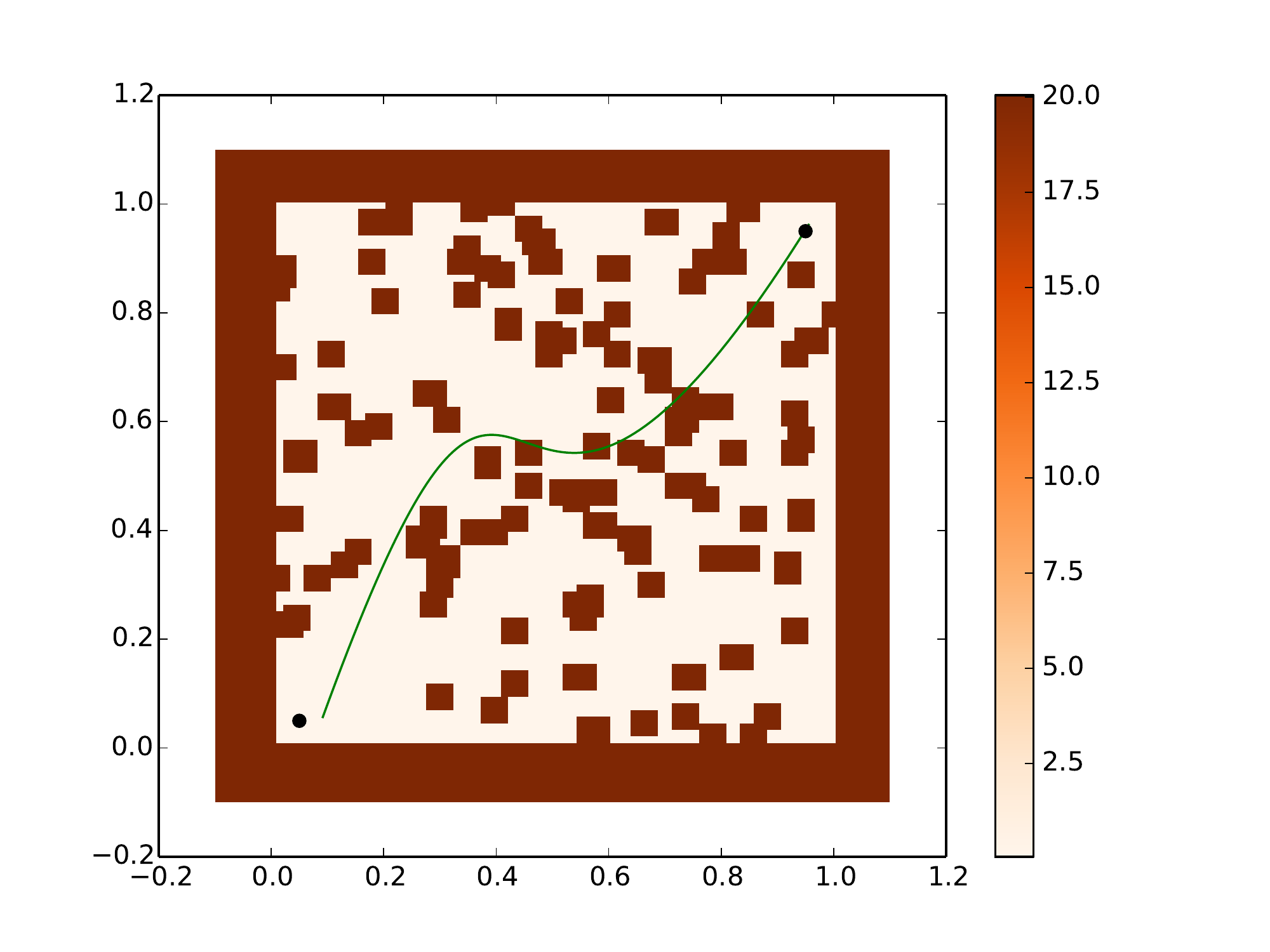}
  \caption{An example trajectory found by EBO.}
  \label{fig:rover_example}
  \end{figure}

 We illustrate the problem in Fig.~\ref{fig:rover_example} with an example trajectory found by EBO. We set the trajectory cost to be $-20.0$ for any collision,  $\lambda$ to be $-10.0$ and the constant $b=5.0$. This reward function is non smooth, discontinuous, and concave over the first two and last two dimensions of the input. These 4 dimensions represent the start and goal position of the trajectory. 
 We maximize the reward function $f$ over the points on the trajectory. All the methods choose a batch of $500$ trajectories to evaluate. Each method is initialized with $10^4$ trajectories randomly uniformly selected from $[0,1]^{60}$ and their reward function values. %
 We again compare EBO with BO-SVI, BO-Add-SVI and CEM~\citep{szita2006learning}. All the methods choose a batch of $500$ trajectories to evaluate. Each method is initialized with $10^4$ trajectories randomly uniformly selected from $[0,1]^{60}$ and their reward function values. The initializations are the same for each method, and we repeat the experiments $5$ times. CEM uses the top $30\%$ of the $10^4$ initial observations to fit its initial Gaussian distribution. At the end of each iteration in CEM, $30\%$ of the new observations with top values were used to fit the new distribution. For all the BO based methods, we use the maximum value of the reward function, $5.0$, in the acquisition function. The standard deviation of the observation noise in the GP models is set to be $0.01$. We set EBO to attempt to have Modrian partitions with fewer than 100 data points, with a hard constraint of no more than 1000 Mondrian partitions. In EBO, we set the hyper parameters $\alpha=1.0$,  $\beta=[2.0,5.0]$, and the Mondrian observation offset $\epsilon=0.01$. In BO-SVI, we used $100$ batchsize in SVI, $200$ inducing points and $500$ iterations to optimize the data likelihood with $0.1$ step rate and $0.9$ momentum. BO-Add-SVI used the same parameters as BO-SVI, except that BO-Add-SVI uses 3 outer loops to randomly select the decomposition parameter $z$ and in each loop, it uses an inner loop of $50$ iterations to maximize the data likelihood over the kernel parameters. The batch BO strategy used in BO-SVI and BO-Add-SVI is identical to the one used in each Mondrian partition of EBO. 

%% file: sup_disc.tex
\section{Discussion}
\subsection{Failure modes of EBO}
EBO is a general framework for running large scale batched BO in high-dimensional spaces. Admittedly,  we made some compromises in our design and implementation to scale up BO to a degree that conventional BO approaches cannot deal with. In the following, we list some limitations and aspects that we can improve in EBO in our future work. 
\begin{itemize}
\item EBO partitions the space into smaller regions $\{[l_j, h_j]\}_{j=1}^J$ and only uses the observations within $[l_j - \epsilon, h_j +\epsilon]$ to do inference and Bayesian optimization. It is hard to determine the value of $\epsilon$. If $\epsilon$ is large, we may have high computational cost for the operations within each region. But if $\epsilon$ is very small, we found that some selected BO points are on the boundaries of the regions, partially because of the large uncertainty on the boundaries.  We used $\epsilon = 0$ in our experiments, but the results can be improved with a more appropriate $\epsilon$.%
\item Because of the additive structure, we need to optimize the acquisition function for each additive component. As a result, EBO has increased computational cost when there are more than 50 additive components, and it becomes harder for EBO to optimize functions more than a few hundred dimensions. One solution is to combine the additive structure with a low dimensional projection approach~\citep{wang2016bayesian}. We can also simply run block coordinate descent on the acquisition function, but it is harder to ensure that the acquisition function is fully optimized.
\end{itemize}

\subsection{Importance of avoiding variance starvation}
Neural networks have been applied in many applications and received success for tasks including regression and classification. While researchers are still working on the theoretical understanding, one hyoothesis is that neural networks ``overfit''~\cite{zhang2016understanding}. Due to the similarity between the test and training set in the reported experiments in, for example, the computer vision community, overfitting may seem to be less of a problem. However, in active learning (e.g. Bayesian optimization), we do not have a ``test set''. We require the model to generalize well across the search space, and using the classic neural network may be detrimental to the data selection process, because of variance starvation (see Section~2). Gaussian processes, on the contrary, are good at estimating confidence bounds and avoid overfitting. However, the scaling of Gaussian processes is hard in general. We would like to reinforce the awareness about the importance of estimating confidence of the model predictions on new queries, i.e., avoiding variance starvation. 

\subsection{Future directions}
Possible future directions include analyzing theoretically what should be the best input space partition strategy, batch worker budget distribution strategy, better ways of predicting variance in a principled way (not necessarily GP), better ways of doing small scale BO and how to adapt it to large scale BO. Moreover, add-GP is only one way of reducing the function space, and there could be others suitable ones too.